\documentclass[10pt,twocolumn,letterpaper]{article}

\usepackage[final]{cvpr}

\usepackage{times}
\usepackage{epsfig}
\usepackage{graphicx}
\usepackage{amsmath}
\usepackage{multirow}
\usepackage{booktabs}
\usepackage{amssymb}
\usepackage{arydshln}
\usepackage{float}
\usepackage[font=small,skip=5pt]{caption}
\usepackage{dsfont}
\usepackage{pifont}%

\newcommand{\xmark}{\ding{55}}

\usepackage{color}
\definecolor{rowblue}{RGB}{220,230,240}
\definecolor{myorchid}{RGB}{150,10,30}
\definecolor{myblue}{RGB}{10,30,250}
\definecolor{mygreen}{RGB}{0,200,10}
\definecolor{mypurple}{RGB}{120,0,120}
\definecolor{mymaroon}{RGB}{128,0,0}
\definecolor{myorange}{RGB}{255,69,0}
\definecolor{myteal}{RGB}{20,150,200}

\newcommand{\myparagraph}[1]{\vspace{1mm} \noindent \textbf{#1}}

\usepackage[pagebackref=true,breaklinks=true,colorlinks,bookmarks=false]{hyperref}

\def\cvprPaperID{9642} %
\def\confName{CVPR}
\def\confYear{2022}

\begin{document}

\title{Robust Cross-Modal Representation Learning with Progressive Self-Distillation}

\author{Alex Andonian\thanks{This work was done when the author was an intern at Amazon.}\\
MIT CSAIL\\
{\tt\small andonian@mit.edu}
\and
Shixing Chen, Raffay Hamid\\
Amazon Prime Video\\
{\tt\small \{shixic, raffay\}@amazon.com}
}

\maketitle

\begin{abstract}

\noindent The learning objective of vision-language approach of CLIP~\cite{radford2021learning} does not effectively account for the noisy many-to-many correspondences found in web-harvested image captioning datasets, which contributes to its compute and data inefficiency. To address this challenge, we introduce a novel training framework based on cross-modal contrastive learning that uses progressive self-distillation and soft image-text alignments to more efficiently learn robust representations from noisy data. Our model distills its own knowledge to dynamically generate soft-alignment targets for a subset of images and captions in every minibatch, which are then used to update its parameters. Extensive evaluation across $14$ benchmark datasets shows that our method consistently outperforms its CLIP counterpart in multiple settings, including: (a) zero-shot classification, (b) linear probe transfer, and (c) image-text retrieval, without incurring extra computational cost. Analysis using an ImageNet-based robustness test-bed~\cite{taori2020measuring} reveals that our method offers better effective robustness to natural distribution shifts compared to both ImageNet-trained models and CLIP itself. Lastly, pretraining with datasets spanning two orders of magnitude in size shows that our improvements over CLIP tend to scale with number of training examples. \vspace{-0.4cm}

\end{abstract}

\section{Introduction}

\noindent The convergence of self-supervised pretraining techniques in natural language processing and computer vision have brought about a renaissance of cross-modal representation learning methods~\cite{radford2021learning,furst2021cloob,shen2021much,algan2021metalabelnet,li2021supervision,wang2021efficientclip,morgado2021audio,jia2021scaling} where large-scale weakly correlated multimodal data (\textit{e.g.,} image-text pairs) is used to learn cross-modal representations using contrastive learning techniques. In particular, the recently proposed CLIP~\cite{radford2021learning} model has garnered significant attention due to its impressive zero-shot recognition ability and excellent transfer performance on downstream tasks.

\begin{figure}[t!]
    \centering
    \includegraphics[width=1.0\linewidth]{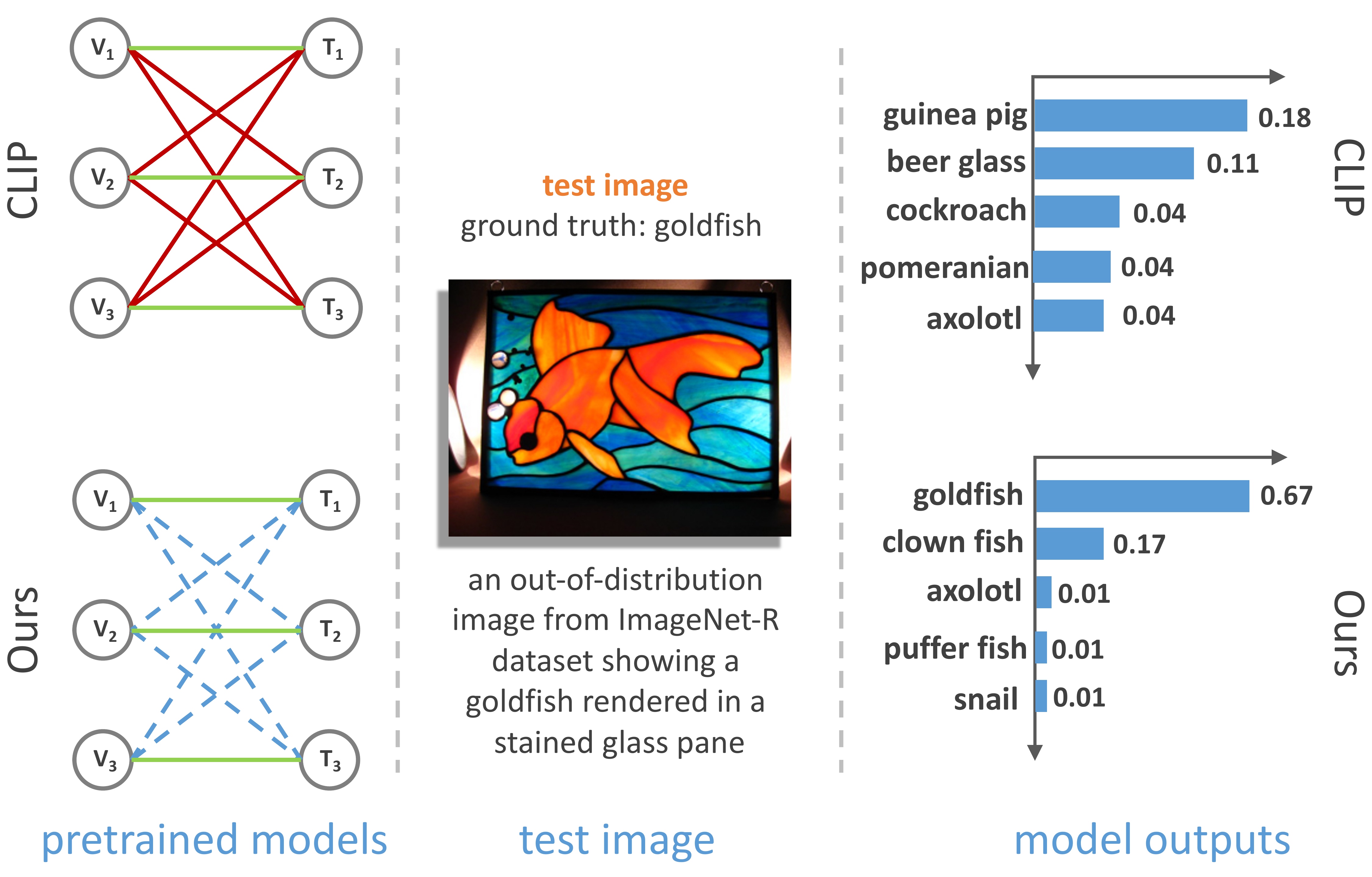}
    \caption{
    \textbf{Illustrative Comparison with CLIP -- } CLIP~\cite{radford2021learning} learns a joint vision-language embedding space by bringing corresponding image-text representation together (green links), while repelling unpaired instances away from each other (red links). This formulation does not account for potential semantic similarity between negative samples.
    We address this issue by learning to predict a distribution of soft-alignment targets (dotted blue edges) in a given minibatch, thereby allowing our model to learn more robust representations.
    This robustness is evident when comparing predicted distributions on an out-of-distribution image from ImageNet-R dataset~\cite{hendrycks2021many}, where unlike CLIP, our method can correctly classify a goldfish rendered in a stained glass pane.
    }\vspace{-0.45cm}
    \label{fig:teaser}
\end{figure}

However, despite their recent success, multimodal pretraining methods like CLIP~\cite{radford2021learning} are data and compute inefficient. Much of CLIP's success can be attributed to its voracious appetite for training data, utilizing $400\textrm{M}$ image-text pairs and an estimated $3,584$ GPU days for pretraining. As the scale of data increases, pretraining requirements of these methods become increasingly expensive, thereby limiting their widespread adoption in a sustainable manner.

This data and compute inefficiency of CLIP~\cite{radford2021learning} can be partially attributed to the underlying assumptions it makes about the web-harvested data it uses for training. Several mainstream vision-language datasets utilize the alt-text HTML attribute of images scraped from archived web pages~\cite{sharma2018conceptual,thomee2016yfcc100m,changpinyo2021conceptual} where captions can often have words unrelated to their corresponding image-content~\cite{sharma2018conceptual}. However, CLIP~\cite{radford2021learning} models the caption for each image to be accurately and exclusively related to only that image (see Figure \ref{fig:teaser}). Moreover, when using larger batch sizes ($32\textrm{K}$ used for CLIP), the likelihood of observing negatives with high semantic similarity increases which can further degrade the learned representations especially those associated with shared semantics between faulty negatives~\cite{arora2019theoretical}. 

To address this challenge, we propose to model the many-to-many relationships between images of web-harvested datasets and their corresponding captions more accurately using \textit{soft probabilities} rather than hard pairing labels. Specifically, we propose a simple yet effective framework for robust contrastive language-image pretraining that uses progressive self-distillation and soft image-text alignment targets to more efficiently learn from noisy data.
Instead of explicitly finding, correcting or even pruning noisy correspondences~\cite{wang2021efficientclip,zolfaghari2021crossclr}, our joint student-teacher model dynamically generates a new set of soft-alignments for a random subset of images and captions in every minibatch.
This enables our method to model many-to-many relationships while simultaneously re-calibrating potentially poorly matched instances without needing to identify them. Over the course of training, our network generates soft-alignments for increasingly large subsets of a minibatch, effectively becoming its own teacher. We identify several key elements that allow the student network to predict its targets without representation collapse or reinforcing its mistakes.

We use multiple pretraining datasets to extensively compare our approach to CLIP~\cite{radford2021learning} evaluated on $14$ benchmark datasets, where our approach consistently outperforms CLIP under multiple settings. Analysis using an ImageNet-based robustness test-bed~\cite{taori2020measuring} shows that our method offers better effective robustness to natural distribution shifts compared to both ImageNet-trained models as well as CLIP. Pretraining with datasets spanning two orders of magnitude in size shows that our improvements over CLIP tend to scale with number of training examples. Lastly, the simplicity of our approach allows it to be readily incorporated into existing and future methods.

\section{Related Work}

\myparagraph{a. Self-Supervised Representation Learning:}
Self-supervised learning (SSL) approaches use a \textit{pretext task} to automatically generate a supervision signal from the data itself, thereby eliminating the dependence on expensive manual data-labeling~\cite{jing2020self}. Pretext tasks in computer vision include spatial reasoning~\cite{noroozi2016unsupervised,doersch2015unsupervised,kim2019self,pathak2016context,gidaris2018unsupervised}, temporal context~\cite{misra2016shuffle,lee2017unsupervised,kim2019self,mobahi2009deep,han2019video}, and other visual properties such as hue~\cite{deshpande2015learning,larsson2016learning,zhang2016colorful,zhang2017split}, brightness~\cite{jason2016back,tian2020cmc} or optical flow~\cite{jason2016back,tian2020contrastive,piergiovanni2020evolving,yin2018geonet}, reconstruction of modified inputs~\cite{vincent2008extracting,pathak2016context,zhang2016colorful}, and classifying inputs with pseudo-labels~\cite{doersch2015unsupervised,dosovitskiy2015discriminative,pathak2017learning} or pseudo-clusters~\cite{caron2018deep,caron2020unsupervised,zhuang2019local,zhuang2020unsupervised}. A promising subset of SSL methods uses a variant of instance discrimination framework~\cite{wu2018unsupervised,dosovitskiy2014discriminative} which learns to align augmented versions of features while distinguishing them from features of other instances using a contrastive loss~\cite{caron2020unsupervised,he2019moco,misra2020self,chen2020simple}.

\myparagraph{b. Vision-Language Pretraining:} Joint vision-language pretraining (VLP) is an active research area~\cite{radford2021learning,furst2021cloob,shen2021much,algan2021metalabelnet,li2021supervision} where the availability of large-scale image-text datasets \textit{e.g.,} YFCC$100$M~\cite{thomee2016yfcc100m} and Conceptual Captions~\cite{sharma2018conceptual,changpinyo2021conceptual} has played a key role in its progress. Although multiple concurrent works are being proposed to further improve VLP models~\cite{wang2021efficientclip}, our work is different from them in a few important ways. Specifically, unlike EfficientCLIP~\cite{wang2021efficientclip} that proposes an ensemble approach to obtain a less noisy data subset for cross-modal training, our method attempts to side-step this problem altogether by re-purposing as opposed to completely removing noisy data. Similarly, DeCLIP~\cite{li2021supervision} improves on the data-efficiency of CLIP~\cite{radford2021learning} by leveraging intra-model contrastive learning along with a nearest-neighbor feature bank to augment negatives. However, incorporating these supervision sources can be computationally expensive. In contrast, our approach offers a simple yet effective way to improve the data-efficiency of CLIP~\cite{radford2021learning} without incurring additional computational cost.

\myparagraph{c. Learning from Noisy Data:}
Several techniques have been developed to increase the robustness of label noise especially in the supervised context~\cite{reed2014training,zhang2018generalized,patrini2017making,han2018co,li2017learning}. These techniques include loss functions that reduce impact of outliers~\cite{zhang2018generalized,wang2019symmetric}, 
metalearning procedures that learn how to correct sources of label noise~\cite{li2019learning,shu2019meta,zhang2020distilling,algan2021metalabelnet},
loss correction approaches that model label noise~\cite{reed2014training,morgado2020avid,morgado2021robust},
regularization techniques aimed at lowering the impact of noise~\cite{pereyra2017regularizing}, and noise filtering processes that iteratively refine dataset labels and retrain models to obtain a more robust final model~\cite{northcutt2021confident}.
However, these works investigate noise-robust methods in the context of common object detection and classification tasks and cannot directly be applied effectively to cross-modal pretraining tasks.
As it stands, noise-robust VLP pretraining methods are still a relatively unexplored topic.

\myparagraph{d. Knowledge Distillation:}
Approaches for knowledge distillation (KD)~\cite{hinton2015distilling} aim to transfer knowledge from one model (\textit{i.e.}, a teacher) to another (\textit{i.e.}, a student). While KD techniques are often motivated by certain performance and efficiency goals~\cite{bucilua2006model,chen2017learning,romero2014fitnets,li2020few}, researchers have also found that KD methods serve as an effective regularization technique that can reduce model overfitting and improve generalization capabilities ~\cite{muller2019does,li2017learning,ding2019adaptive,liu2021inflate}.
Our approach is motivated by the recent success of self-knowledge distillation approaches ~\cite{hahn2019self,xu2019data,yun2020regularizing} that use the student network as a teacher under supervised settings to achieve high accuracy with reduced computational cost. To the best of our knowledge, we are among the first to investigate progressive self-distillation in the context of vision-language pretraining.
\vspace{-0.5em}

\newcommand{\imencoder}{\ensuremath{f_{v}} }
\newcommand{\textencoder}{\ensuremath{f_{t}} }
\newcommand{\timencoder}{\ensuremath{\tilde{f}_{v}} }
\newcommand{\ttextencoder}{\ensuremath{\tilde{f}_{t}} }
\newcommand{\vi}{\ensuremath{\mathbf{v}_i}}
\newcommand{\ti}{\ensuremath{\mathbf{t}_i}}
\newcommand{\vj}{\ensuremath{\mathbf{v}_j}}
\newcommand{\tj}{\ensuremath{\mathbf{t}_j}}
\newcommand{\vti}{\ensuremath{\mathbf{\tilde{v}}_i}}
\newcommand{\tti}{\ensuremath{\mathbf{\tilde{t}}_i}}
\newcommand{\vtj}{\ensuremath{\mathbf{\tilde{v}}_j}}
\newcommand{\ttj}{\ensuremath{\mathbf{\tilde{t}}_j}}
\newcommand{\tk}{\ensuremath{\mathbf{t}_k}}
\newcommand{\zi}{\ensuremath{\mathbf{z}_i}}
\newcommand{\zj}{\ensuremath{\mathbf{z}_j}}
\newcommand{\zvi}{\ensuremath{\mathbf{z}^v_i}}
\newcommand{\zti}{\ensuremath{\mathbf{z}^t_i}}
\newcommand{\zvj}{\ensuremath{\mathbf{z}^v_j}}
\newcommand{\zvk}{\ensuremath{\mathbf{z}^v_k}}
\newcommand{\ztj}{\ensuremath{\mathbf{z}^t_j}}
\newcommand{\ztk}{\ensuremath{\mathbf{z}^t_k}}
\newcommand{\Linfonce}{\ensuremath{\mathcal{L}_{\text{InfoNCE}}}}
\newcommand{\Lv}{\ensuremath{\mathcal{L}_{v}}}
\newcommand{\Lt}{\ensuremath{\mathcal{L}_{t}}}
\newcommand{\Lvt}{\ensuremath{\widetilde{\mathcal{L}_{v}}}}
\newcommand{\Ltt}{\ensuremath{\widetilde{\mathcal{L}_{t}}}}
\newcommand{\V}{\ensuremath{\mathbf{V}}}
\newcommand{\T}{\ensuremath{\mathbf{T}}}
\newcommand{\Vt}{\ensuremath{\mathbf{\tilde{V}}}}
\newcommand{\Tt}{\ensuremath{\mathbf{\tilde{T}}}}
\newcommand{\Av}{\ensuremath{\mathbf{A}^v}}
\newcommand{\At}{\ensuremath{\mathbf{A}^t}}
\newcommand{\taut}{\ensuremath{\tilde{\tau}}}
\newcommand{\Irm}{\ensuremath{\textrm{I}}}
\newcommand{\Nrm}{\ensuremath{\textrm{N}}}
\newcommand{\Vta}{\ensuremath{\mathbf{\tilde{V}}^a}}
\newcommand{\Tta}{\ensuremath{\mathbf{\tilde{T}}^a}}
\newcommand{\Vu}{\ensuremath{\mathbf{V}^u}}
\newcommand{\Tu}{\ensuremath{\mathbf{T}^u}}

\vspace{-0.5em}
\section{Methods}
\vspace{-0.5em}

\begin{figure*}[t!]
    \centering
    \includegraphics[width=0.825\linewidth]{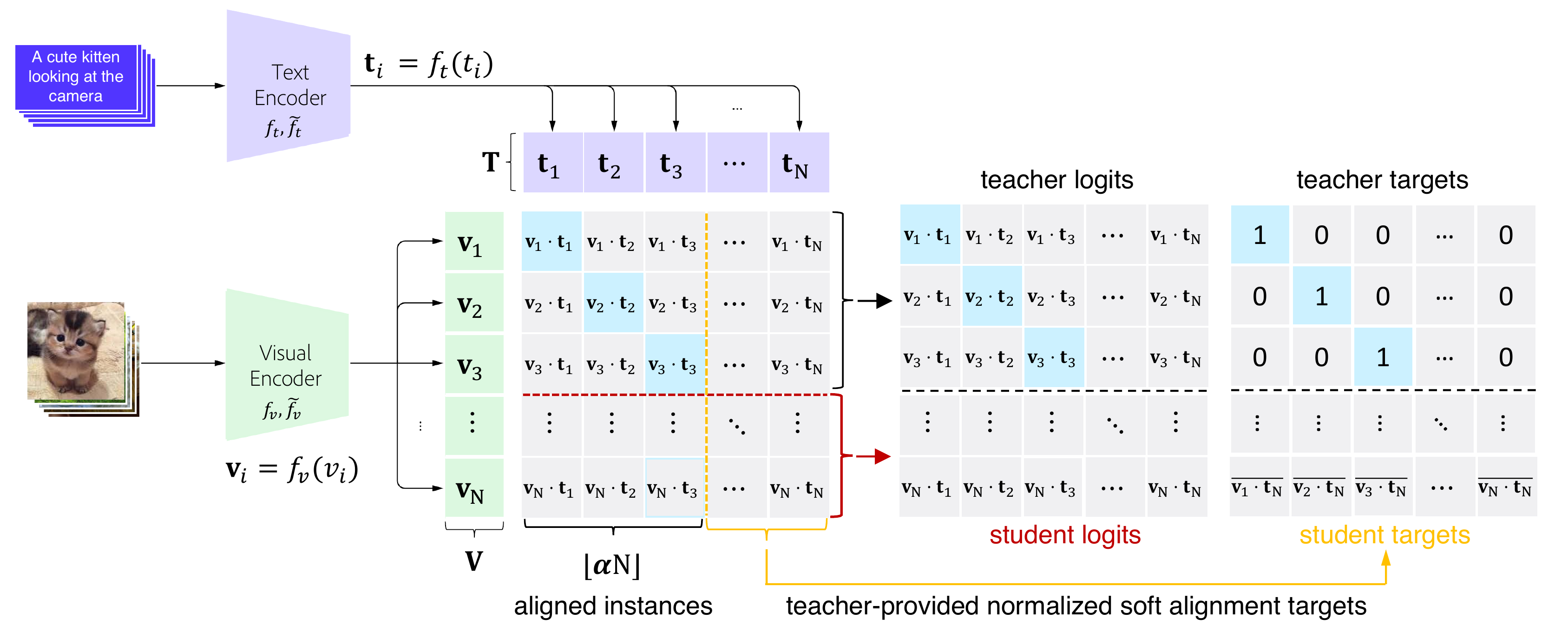}
    \caption{
    \small
\textbf{Approach Overview -- }We partition a minibatch into aligned instances (rows above the dotted red line) and unaligned instances (rows below the dotted red line).
    Our teacher network is trained on the aligned instances using the standard InfoNCE loss. By normalizing the predictions from the opposite modality (columns to the right of the dotted yellow line), the teacher provides the estimated soft-alignments for the unaligned data to supervise the student. Over the course of training, the proportion of soft-alignments provided to the student increases as the teacher's representations become more reliable.
    \vspace{-0.3cm}
    }
    \label{fig:method}
\end{figure*}

\noindent Unlike CLIP~\cite{radford2021learning}, we view the problem of learning aligned vision-language representations from web-scale weakly-annotated data as the challenge of learning many-to-many relationships from noisy image-text correspondences. To address this challenge, we propose a novel vision-language pretraining method that progressively distills a model's own knowledge to soften its initially hard-target alignments, thereby enabling it to learn more transferable representations from the same amount of training data (see Figure~\ref{fig:method}). In the following, we first establish cross-modal contrastive learning objective and identify some of its limitations. We then introduce our novel progressive self-distillation approach and explain how it addresses these limitations.

\subsection{Preliminaries}
\vspace{-0.2em}
\noindent We consider a batch of $\textrm{N}$ semantically paired image-text tuples $\{(v_i, t_i )\}_{i=1:\textrm{N}}$ drawn from a cross-modal dataset.
The goal of cross-modal contrastive pretraining is to learn encoders \imencoder for image data and \textencoder for text data such that for a given semantically related instance $(v_i, t_i)$, the encoded $\ell_2$-normalized embeddings $\vi = \imencoder(v_i)$ and $\ti = \textencoder(t_i)$ with $\vi,\ti \in \mathbb{R}^d$ are close together (\textit{i.e.}, ``aligned'') under some distance metric, while the unpaired image and text embeddings are farther apart (\textit{i.e.}, ``unaligned'').

\subsection{Contrastive Learning with InfoNCE Loss}
\vspace{-0.2em}
\noindent Recall that CLIP~\cite{radford2021learning} trains these image and text encoders with a contrastive loss by minimizing the InfoNCE\cite{oord2018representation} loss $\Linfonce = \Lv + \Lt$, where $\Lv$ is the loss for aligning images to text and $\Lt$ is the loss for aligning text to images. Specifically, $\Lv$ is defined as:
\begin{align}
    &\Lv = - \frac{1}{\Nrm}\sum_{i=1}^\Nrm\sum_{j=1}^\Nrm \Irm_{ij} \log \textrm{P}_v(\vi, \tj; \tau)\\
    &\textrm{P}_v(\vi, \tj;\tau) = \frac{\exp(\text{sim}(\vi, \tj)/\tau)}{\sum_{k=1}^\Nrm \exp(\text{sim}(\vi, \tk)/\tau)}
\end{align}
where $\text{sim}(\vi, \tj) = \vi^T\tj$ is typically chosen to be the dot product (cosine similarity), $\tau$ is a learnable softmax temperature parameter and $\textrm{I}_{ij}$ is an element from the identity matrix $\mathbf{I}_{\textrm{N}}$. Since InfoNCE is symmetric,  $\Lt$ and $\textrm{P}_t$ are defined in a similar manner.

For convenience, let $\V, \T \in \mathbb{R}^{\textrm{N} \times d}$ be matrices that contain a batch of image and text embeddings whose rows are populated with $\mathbf{v}_{1:\textrm{N}}$ and $\mathbf{t}_{1:\textrm{N}}$, respectively. Then, the InfoNCE loss can be re-written compactly in matrix form as:
\begin{equation}
    \Linfonce = \mathcal{H}(\mathbf{I}_\textrm{N}, \rho(\V{\T}^\top)) + \mathcal{H}(\mathbf{I}_\textrm{N}, \rho(\T{\V}^\top)),
    \label{eqn:infonce}
\end{equation}
where $\mathcal{H}$ is the batched (row-wise) cross-entropy function with mean reduction and $\rho$ is the standard softmax function applied row-wise such that each row sums to one.

Equation \ref{eqn:infonce} shows that InfoNCE loss is simply the cross entropy between a one-hot distribution $\Irm_{ij}$ and estimated probability $\textrm{P}_{v}(\vi, \tj; \tau)$. It enforces the strict assumption that an image $v_i$ selected from a batch should be paired exclusively with text $t_i$ while being repelled from all other $t_j$.

However, this assumption generally does not hold for two important reasons. First, it is likely that a given image would be aligned to several text captions to different degrees especially under large batch-size settings. Second, the ground truth pairings in large-scale weakly-annotated datasets may be simply incorrect or describe loose correlations between images and their corresponding texts.

\subsection{Distillation through Soft-Alignments}
\vspace{-0.2em}
\noindent To address the aforementioned limitations of using the InfoNCE loss to train on noisy cross-modal data, we propose to adopt a knowledge distillation framework where the predictive probabilities produced by a teacher network are used as soft target distributions to train a student network. 

Our framework offers two key advantages.
First, in the process of generating target image-text alignments, a well-trained teacher can combat poorly captioned images by re-pairing them with stronger semantic matches from the batch, thereby providing a cleaner learning signal for the student network.
Second, by providing soft targets, the teacher can convey many-to-many relationships in a batch.

Specifically, to estimate the correspondence between image $v_i$ and text $t_i$, our teacher model employs image and text encoders \timencoder and \ttextencoder to compute $\ell_2$-normalized teacher embeddings $\vti$ and $\ttj$ which are similarly batched as rows in the matrices $\Vt$ and $\Tt$ respectively.
Our method uses a swapped prediction strategy to produce soft target distributions $\Av$ and $\At$ to supervise student training. These target distributions are defined as:
\begin{equation}
\Av = \rho(\Tt\Vt^\top; \taut) \text{ and } \At =  \rho(\Vt \Tt^\top; \taut)
\end{equation}
where $\rho$ is the standard softmax function now using a secondary teacher temperature $\taut$ that transforms and re-scales raw logits into probabilities. 

Swapped prediction improves on the well established forward bootstrapping approach~\cite{reed2014training} by using predictions from opposite modality. Unlike bootstrapping, swapped prediction computes the image alignment scores $\Av$ from the text encoder posterior probabilities and vice-versa, thus aggregating information over all other instances from the opposite modality.
Intuitively, the strength of alignment from image $v_i$ to text $t_j$ is based on the probability that text $t_j$ should be matched with image $v_i$ compared to all other $v_j$. This strategy has shown promise in related contrastive learning settings~\cite{morgado2021robust}, which is consistent with our results.

Replacing the $\mathbf{\textrm{I}}_{\textrm{N}}$ targets in Equation~\ref{eqn:infonce} with estimated soft-alignment probabilities $\Av$ and  $\At$ allows the teacher to re-calibrate the attractive and repulsive forces between image and text embeddings in the representation space based on its estimated similarity between instances.
For instance, a faulty negative pair $(v_i, t_j)$, which may have high semantic similarity is assigned a similarity score of zero by the InfoNCE loss, whereas our method provides $\textrm{A}_{ij}$ as a target which should be larger given a well trained teacher.

\subsection{Progressive Self-Distillation}
\vspace{-0.2em}
\noindent We now explain how to start such a teacher network and how its contribution to learning process evolves over time.
\vspace{-2mm}
\subsubsection{Teacher Network Selection}
\vspace{-2mm}
Conventional KD and SSL methods offer numerous potential teacher choices, \textit{e.g.,} larger but static pretrained teacher networks
\cite{hinton2015distilling}, or networks that share the same model architecture but use weights from a previous epoch~\cite{lin2021self}, or as an exponential moving average~\cite{he2019moco}. The primary drawback to these approaches is reduced computational and memory efficiency as they require a secondary inference stage using additional model weights that must be kept in memory.

To circumvent these issues, we adopt a \emph{self-distillation} strategy where the student network acts as its own teacher (\emph{i.e.,} $\imencoder=\timencoder, \textencoder=\ttextencoder$). 
The idea here is to update the targets of the student contrastive objective using the current state of the model.
Intuitively, as the learning improves over time, its representation can be trusted to make more accurate predictions.
This mitigates the negative effects of noisy pairings as incorrect pairs are increasingly likely to be inconsistent with the consensus learned from the rest of the data as training progresses.
By refining inconsistent alignments, the model can develop more coherent representations, which further improves its ability to evaluate the consistency of noisy image-text pairs.
\vspace{-2mm}
\subsubsection{Progressing from Student to Teacher}
\vspace{-2mm}
\noindent As our objective relies on some basic level of alignment between corresponding image and text representations, we introduce a novel procedure that progressively increases the contribution of self-distillation to the contrastive learning process over the course of training.
Our model therefore dynamically evolves into its own teacher as training progresses, which differs from the standard knowledge distillation setting where the teacher is often static and separate.

We achieve this dynamic progression by randomly partitioning a batch of $\Nrm$ image-text pairs into $\Nrm^a = \lfloor \alpha \Nrm\rfloor$ ``aligned'' instances and $\Nrm^u = \Nrm-\lfloor \alpha \Nrm \rfloor$ ``unaligned'' instances where  $\alpha \in [0, 1]$ determines their relative proportions. 
The aligned subset is used to train the teacher network using the hard ground truth pairings and the standard InfoNCE loss.
The teacher network then employs our aforementioned swapped prediction strategy to estimate soft-alignments on the unaligned instances to supervise the student.
We refer to this random minibatch partitioning as \emph{dynamic} as opposed to static as the global partitioning of instances into aligned and unaligned subsets is refreshed for each training epoch.

To increase the strength of the teacher's influence on learning, we decrease the value of $\alpha$ gradually, in the same way that the learning rate can be scheduled. While there are several strategies to decrease $\alpha$ as a function of the training iteration, \emph{e.g.,} step-wise, linear, \textit{etc.}, we use a cosine-annealing schedule~\cite{li2020few} specified by a start and end value.

To summarize our overall learning procedure, we first compute the batched student-teacher embedding with $\V =\Vt$ and $\T=\Tt$. Next, we extract the first $\Nrm^a$ rows to form aligned subset of teacher embeddings $\Vta,\Tta$, and the last $\Nrm^u$ rows for the unaligned student embeddings $\Vu,\Tu$. Altogether, our final objective function is defined as:
\vspace{-1mm}
\begin{align*}
    \Linfonce^{\textrm{PSD}} = \alpha &\left[
        \mathcal{H}(\mathbf{I}_{\Nrm_a}, \rho(\Vta{\Tt}^\top)) +
        \mathcal{H}(\mathbf{I}_{\Nrm_a}, \rho(\Tta{\Vt}^\top))
    \right] + \\
    (1 - \alpha)&\left[
        \mathcal{H}(\mathbf{A}^v, \rho(\Vu{\T}^\top)) +
        \mathcal{H}(\mathbf{A}^t, \rho(\Tu{\V}^\top))
    \right]
    \label{eqn:infonce}
\end{align*}
where $\mathbf{I}_{\Nrm_a} \in \mathbb{R}^{\Nrm_a \times \Nrm}$ is the zero-padded identity matrix, while $\Av,\At$ are indexed to match the unaligned student embeddings.

\section{Experiments}
\begin{table*}[!h]
	\small{
    \centering
    \begin{tabular}{ccccccccccccl}
    \shortstack[c]{Pretraining\\ Dataset} &
    \shortstack[c]{Method} &
    \rotatebox{70}{\footnotesize{Cifar10\cite{krizhevsky2009learning}}} &
    \rotatebox{70}{\footnotesize{Cifar100\cite{krizhevsky2009learning}}} &
    \rotatebox{70}{\footnotesize{Caltech101\cite{fei2006one}}} &
    \rotatebox{70}{\footnotesize{Places365\cite{zhou2017places}}} &
    \rotatebox{70}{\footnotesize{ObjectNet\cite{barbu2019objectnet}}} &
    \rotatebox{70}{\footnotesize{ImageNet-R\cite{hendrycks2021many}}} &
    \rotatebox{70}{\footnotesize{ImageNet-O\cite{hendrycks2019natural}}} &
    \rotatebox{70}{\footnotesize{Imagenet-A\cite{hendrycks2019natural}}} &
    \rotatebox{70}{\footnotesize{ImageNetV2\cite{recht2019imagenet}}} &
    \rotatebox{70}{\footnotesize{ImageNet\cite{deng2009imagenet}}} &
    \rotatebox{70}{\footnotesize{Average}} \\
    \midrule
    \multirow{2}{*}{COCO}     & CLIP & 64.14 & 19.57 & 32.88 & 12.78 & 4.98 & 8.27  & 8.0  & 3.32 & 7.41 & 8.18 & 16.87 \\
                              & Ours & \textbf{66.74} & \textbf{24.49} & \textbf{34.26} & \textbf{14.15} & \textbf{6.18 }& \textbf{11.25} & \textbf{9.85 }& \textbf{5.32 }& \textbf{8.99 }& \textbf{9.49 }& $\textbf{19.07}^{\textbf{\textcolor{mygreen}{+2.22}}}$ \\
    \cdashline{1-13}
    \multirow{2}{*}{CC3M}     & CLIP & 73.90 & 30.60 & 54.07 & 24.54 & 4.49 &28.33 & 18.5 & 7.827 & 21.43 & 23.56 & 28.73 \\
                              & Ours & \textbf{80.15} & \textbf{38.27} & \textbf{64.45} & \textbf{28.07} & \textbf{9.21} &\textbf{37.31} & \textbf{26.2} & \textbf{10.81} & \textbf{26.70} & \textbf{27.96} & $\textbf{34.91}^{\textbf{\textcolor{mygreen}{+6.19}}}$ \\
    \cdashline{1-13}
    \multirow{2}{*}{CC12M}     & CLIP & 75.29 & 41.94 & 75.86 & 31.29 & 12.50 & 51.34    & 31.90 & 13.25 & 34.89 & 37.87 & 40.61 \\
                               & Ours & \textbf{84.84} & \textbf{51.34} & \textbf{80.00} & \textbf{34.08} & \textbf{15.24} & \textbf{59.29} & \textbf{33.0} & \textbf{18.85} & \textbf{39.16} & \textbf{42.24} & $\textbf{45.85}^{\textbf{\textcolor{mygreen}{+5.23}}}$ \\
    \bottomrule
    \end{tabular}
    \caption{\textbf{Zero-Shot Image Classification Comparison -- } Zero-shot top-1 accuracy (\%) of our method compared to baseline CLIP on numerous ImageNet-based benchmark datasets using different pretraining datasets varying in scale.}
    \label{tab:zeroshot}
	}
\end{table*}

\noindent We start by describing our experimental setup which aims to match CLIP~\cite{radford2021learning} as closely as possible for fair comparison. We then demonstrate the advantages of our method over CLIP~\cite{radford2021learning} for: (a) zero-shot classification, (b) finetuning (\textit{i.e.}, linear probe), and (c) image-text retrieval.

\subsection{Pretraining Datasets}
\noindent
We apply our pretraining method to three image-text datasets varying in scale, scope and noise:

\myparagraph{a. MS COCO Captions~\cite{lin2014microsoft}} -- A widely used standard image captioning benchmark dataset with approximately $118$K images, each labeled with $5$ human evaluated captions, and a testing set of $5$K images.

\myparagraph{b. Conceptual Captions $3$M} (CC$3$M)~\cite{sharma2018conceptual} -- A collection of over $3$M images and their raw descriptions harvested from the alt-text HTML attribute associated with the web-scraped images, therefore representing a wider variety of content styles. After downloading and preprocessing, we utilized about $2.9$M image-text pairs in our experiments.

\myparagraph{c. Conceptual Captions $12$M} (CC$12$M)~\cite{changpinyo2021conceptual} -- By relaxing multiple image and text filters used in CC$3$M~\cite{sharma2018conceptual}, CC$12$M is a less precise but $4$$\times$ larger set of image-text pairs that covers a wider range of visual concepts. Due to unavailable URLs, we utilize about $10$M examples from this dataset.

\subsection{Pretraining Details}
\noindent
In the following experiments, the image encoders follow \texttt{ViT-B/32} vision transformer architecture proposed in CLIP~\cite{radford2021learning}, while the text encoder's transformer-based architecture follows modification proposed in~\cite{radford2021learning}. Image and text features are projected to a shared 512-D space and L2 normalized before participating in contrastive loss.

Models are trained from scratch for $100$ epochs using the Adam optimizer~\cite{kingma2014adam} with weight decay and a cosine annealing learning rate schedule with warmup~\cite{loshchilov2016sgdr}.
As done in \cite{radford2021learning}, the learnable temperature parameter $\tau$ is initialized to $0.07$ and clamped to values less than $100$.
Automatic mixed-precision~\cite{micikevicius2017mixed} training is used to save on memory and achieve minibatch sizes of $4096$.
Input images are randomly cropped and resized to $224$ $\times$ $224$ resolution during pretraining and the maximum length of the text is limited to $77$ tokens via random sub-sequence sampling similar to~\cite{radford2021learning}.
Training is conducted on as many as $8$ Nvidia A-$100$ GPUs with the longest experiments spanning up to several days. The partitioning factor $\alpha$ is decayed from $0.8$ to $0.2$ over the course of training using cosine annealing.
Details regarding hyperparameter values for different datasets and models are provided in the supplementary material.

We compare our approach to a re-implementation of CLIP based on the methods and pseudo-code described in the original paper as well as open source implementations such as OpenCLIP \cite{ilharco_gabriel_2021_5143773}.
The performance of our re-implementation is consistent with that achieved by OpenCLIP and unreleased CLIP models trained on reduced subsets of the $400$M dataset; thus, we retrain a baseline CLIP model for each pretraining dataset to serve as a proxy.

\subsection{Evaluation Details}
\myparagraph{a. Evaluation Protocol} -- We measure zero-shot and linear probe classification performance using \texttt{Top-1} accuracy. For the linear probe experiments,  we train the linear classifier with the L-BFGS optimizer on extracted visual features as described in \cite{radford2021learning}.
We use standard retrieval metrics: recall at rank $\textrm{K}$ (R@$\textrm{K}$, higher is better) and mean rank (MnR, lower is better), to evaluate the retrieval performance of our model.
R@$\textrm{K}$ (Recall at K) calculates the percentage of test samples for which the correct result is found in the top-$\textrm{K}$ retrieved points to the query sample.

\myparagraph{b. Benchmark Datasets} -- We evaluate the zero-shot and linear probe classification performance of proposed approach on a suite of benchmark evaluation datasets, including ImageNet\cite{deng2009imagenet}, Places365\cite{zhou2017places}, ObjectNet\cite{barbu2019objectnet}, and several recent variants of ImageNet aimed at evaluating the robustness of trained models to natural (as opposed to synthetic) distribution shifts~\cite{hendrycks2021many,hendrycks2019natural,recht2019imagenet}.

\myparagraph{c. Prompt Ensembling with Templates} -- Consistent with previous work, we find that our approach benefits from prompt ensembling to augment the original class label for downstream tasks. For fair comparison, we use the same set of prompt templates published in CLIP\cite{radford2021learning}, which generally take on the form of ``a photo of a \texttt{\{label\}}.''

\subsection{Zero-Shot Image Classification}
\noindent
Following the pretraining stage, we evaluate our method on zero-shot image classification achieved through natural language input, and compare to its CLIP counterpart trained under identical settings.

Table \ref{tab:zeroshot} lists the top-1 accuracy (\%) of zero-shot image classification across a suite of benchmark datasets.
Given a fixed amount of pretraining data, our method considerably outperforms its CLIP counterparts in terms of average top-1 accuracy over all datasets, achieving as much as a 6.19\% absolute improvement in the CC3M pretraining regime.
Notably, our method surpasses CLIP on ImageNet and all of the ImageNet variants.
Our method also achieves substantial performance gains on ImageNet-Renditions (ImageNet-R)\cite{hendrycks2021many}, a dataset specifically aimed at evaluating out-of-distribution robustness, with an average performance gain of nearly 8.5\% when considering Conceptual Captions datasets~\cite{sharma2018conceptual,changpinyo2021conceptual}. Figure \ref{fig:teaser} shows how our method is able to provide more accurate predictions for a goldfish \emph{rendered} in a stained glass pane.

\begin{figure}[t!]
    \centering
    \includegraphics[width=0.875\linewidth]{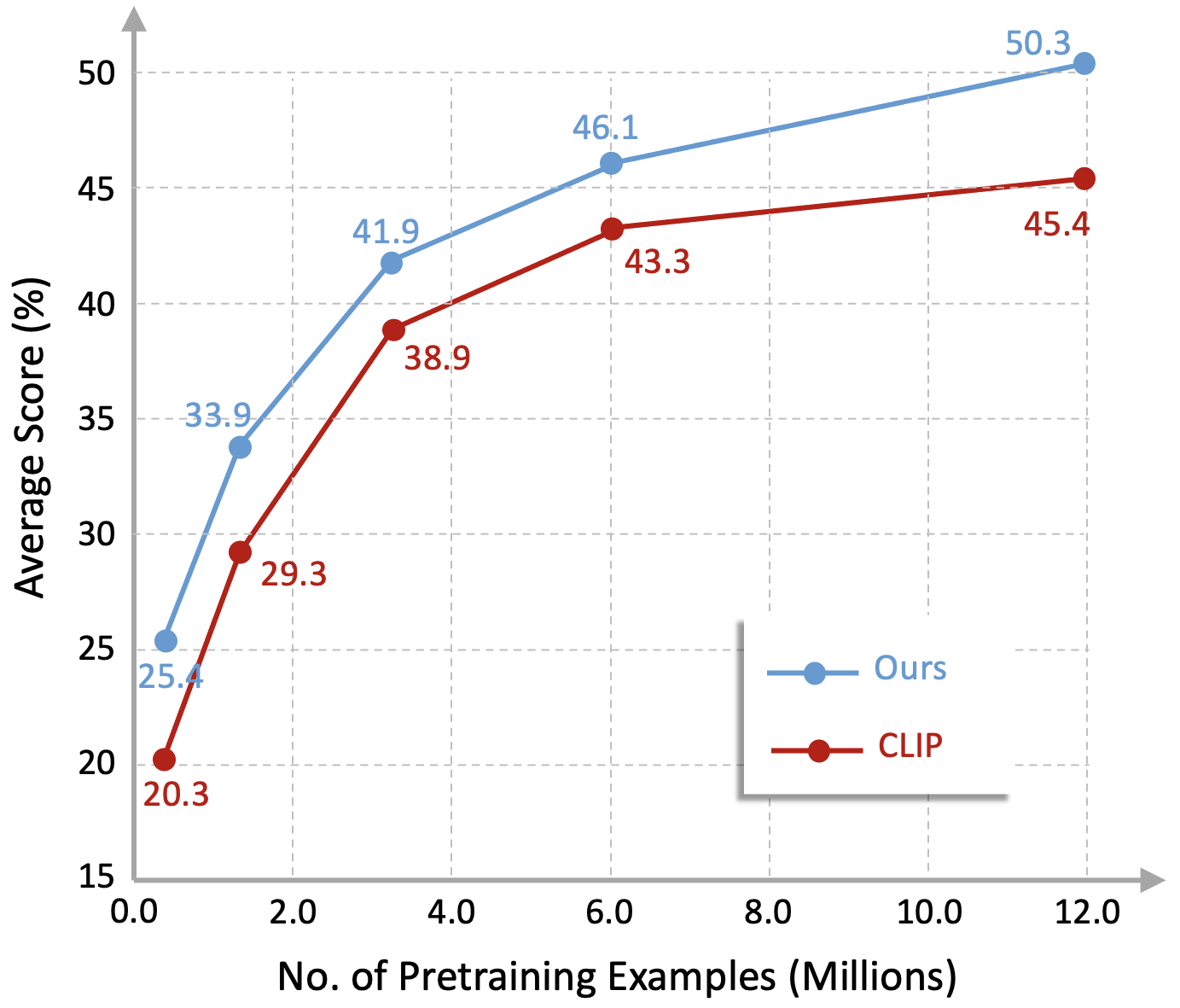}
    \caption{\textbf{Pretraining Data Size and Relative Performance -- }Average zero-shot classification performance of our method compared to CLIP as number of pretraining examples from the CC$12$M dataset grows from $600$K to $12$M.  Our method consistently learns representations more amenable to zero-shot classification than CLIP given same amount of pretraining data.}
    \vspace{-0.3cm}
    \label{fig:data_scaling}
\end{figure}

To better understand the relationship between the amount of pretraining data and relative differences in downstream performance between our method and CLIP, we perform a sweep of experiments that restrict pretraining to $5$, $10$, $25$, $50$, and $100$\% of the CC$12$M dataset. Figure \ref{fig:data_scaling} plots the average zero-shot classification score for both methods, and shows that our method maintains higher performance throughout this range. Interestingly, our method appears to provide the largest performance increases at the boundaries of the plot, suggesting that our method can improve data-efficiency in data-limited regimes, in addition to providing larger performance gains as the dataset size increases.
\begin{figure}[t!]
    \centering
    \includegraphics[width=0.785\linewidth]{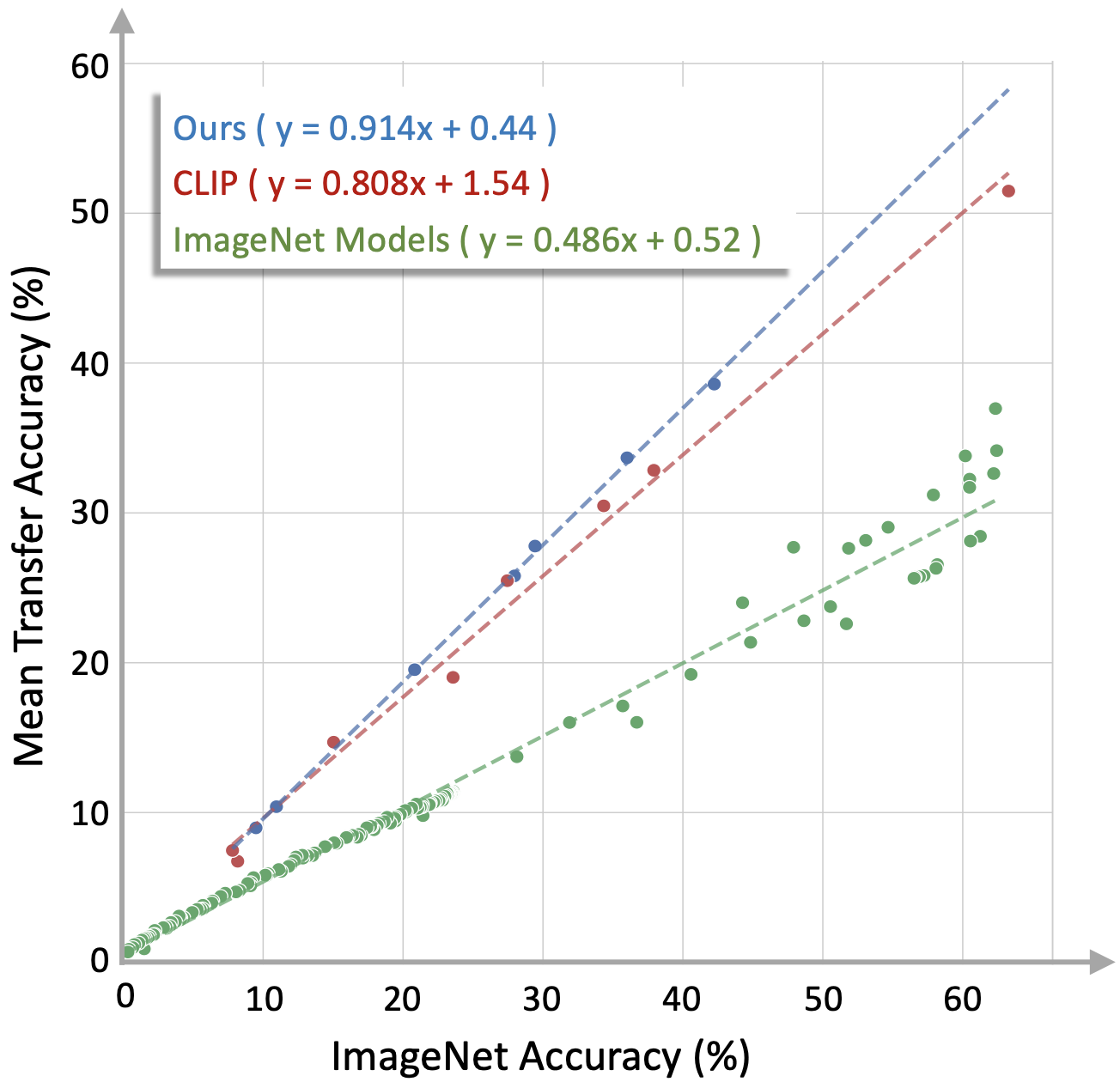}
    \caption{
    \small
    \textbf{Effective Robustness Evaluation -- } Our method produces features more robust to natural distribution shift compared to CLIP features. When comparing models with similar ImageNet performance, our method produces representations that offer better performance on naturally shifted distributions. Mean Transfer Accuracy is computed over the ImageNet-R/O/A/V2 test sets. Best-fit trend lines suggests that the effective robustness of our method (blue) outpaces CLIP (red) as ImageNet accuracy increases.
    }\vspace{-0.3cm}
    \label{fig:robustness}
\end{figure}

\subsection{Evaluation of Effective Robustness}
\noindent
CLIP models~\cite{radford2021learning,taori2020measuring} have been found to be more robust to natural distribution shifts when compared to standard models trained on ImageNet. This phenomena is illustrated in Figure \ref{fig:robustness} with ImageNet accuracy on the x-axis and mean accuracy on ImageNet-A, ImageNet-O, ImageNet-R, and ImageNetV2, on the y-axis.
Previous work~\cite{taori2020measuring,miller2021accuracy} has found that the in-distribution and out-of-distribution accuracies of ImageNet models follow a predictable linear trend (plotted in green), and CLIP models established a trend (plotted in red) of improved effective robustness. Note that the slope of linear-fit to our model is higher than that for the CLIP model, suggesting that our effective robustness improves over CLIP with increasing scale.

\subsection{Linear Probe Performance}
\noindent
We report our linear probe performance on 4 downstream datasets in Table \ref{tab:linear_probe}. Our method outperforms its CLIP counterpart in every case, suggesting that our learned visual features alone (\emph{i.e.,} considering within-modal alignment only) are more transferable than CLIP.

\begin{table}
	\small{
    \centering
    \begin{tabular}{cccccc}
    \small
    \shortstack[c]{Pretraining\\ Dataset} &
    \shortstack[c]{Method} &
    \rotatebox{65}{\footnotesize{Food101~\cite{bossard14}}} &
    \rotatebox{65}{\footnotesize{OxfordPets~\cite{parkhi2012cats}}} &
    \rotatebox{65}{\footnotesize{Birdsnap~\cite{berg-birdsnap-cvpr2014}}} &
    \rotatebox{65}{\footnotesize{ImageNet~\cite{deng2009imagenet}}} \\
    \midrule
    \multirow{2}{*}{COCO}     & CLIP & 53.04          &  76.86         & 37.17          & 52.66 \\
                              & Ours & \textbf{53.60} & \textbf{80.59} & \textbf{42.85} & \textbf{56.21} \\
    \cdashline{1-6}
    \multirow{2}{*}{CC3M}     & CLIP & 53.33          &  78.11         & 37.75          & 56.28 \\
                              & Ours & \textbf{60.69} & \textbf{80.40} & \textbf{43.17} & \textbf{61.00} \\
    \cdashline{1-6}
    \multirow{2}{*}{CC12M}    & CLIP & 67.41          & 85.17          & 41.06          & 59.42 \\
                              & Ours & \textbf{71.87} & \textbf{86.32} & \textbf{47.16} & \textbf{65.33} \\
    \bottomrule
    \end{tabular}
    \caption{\textbf{Linear Probe Performance -- }Linear probe top-1 accuracy (\%) of our method compared to baseline CLIP on four benchmark datasets is given.
     Our method learns visual features that consistently achieve improved finetuning performance, suggesting that our loss helps to improve within-modal feature alignment.}
     \vspace{-0.3cm}
    \label{tab:linear_probe}
	}
\end{table}

\subsection{Image-Text Retrieval}
\noindent Given the cross-modal nature of models under investigation, image-text retrieval consists of two sub-tasks: image-to-text and text-to-image.
We evaluate our method on COCO test set with $5$K images, each with $5$ unique captions.
\begin{table*}[!h]
	\small{
    \centering
    \begin{tabular}{cccccccccc}
    \toprule
    \multirow{2}{*}{\shortstack[c]{Pretraining\\ Dataset}} &
    \multirow{2}{*}{Method} &
    \multicolumn{4}{c}{Text-to-Image} &
    \multicolumn{4}{c}{Image-to-Text} \\
    \cmidrule(lr){3-6} \cmidrule(lr){7-10} 
    && \footnotesize{R@1} $\uparrow$ & \footnotesize{R@5} $\uparrow$ & \footnotesize{R@10} $\uparrow$ & \footnotesize{MnR} $\downarrow$ &
       \footnotesize{R@1} $\uparrow$ & \footnotesize{R@5} $\uparrow$ & \footnotesize{R@10} $\uparrow$ & \footnotesize{MnR} $\downarrow$ \\
    \midrule
    \multirow{2}{*}{COCO}     & CLIP & 27.76          & \textbf{57.34} & \textbf{70.70} & \textbf{23.10} &          27.40 &         56.31  & \textbf{68.65}   & \textbf{20.11} \\
                              & Ours & \textbf{28.42} &          57.14 &          68.86 & 26.11          & \textbf{28.53} & \textbf{56.75} & 68.10            & 22.01 \\
    \cdashline{1-10}
    \multirow{2}{*}{CC3M}     & CLIP & 12.50          & 29.76          & 40.92          & 91.04          & 9.88           & 24.86          &  35.30          & 106.94 \\
                              & Ours & \textbf{16.98} & \textbf{37.12} & \textbf{48.28} & \textbf{63.80} & \textbf{13.19} & \textbf{31.54} & \textbf{43.00}  & \textbf{72.92} \\
    \cdashline{1-10}
    \multirow{2}{*}{CC12M}    & CLIP & 19.64          & 40.66          & 51.72          & 55.23          & 17.63          & 39.67          & 50.77          & 55.29 \\
                              & Ours & \textbf{22.94} & \textbf{46.60} & \textbf{57.82} & \textbf{46.24} & \textbf{22.79} & \textbf{45.95} & \textbf{56.81} & \textbf{43.41} \\
    \midrule 
    \multirow{2}{*}{CC3M}     & CLIP$^{\text{\textasteriskcentered}}$ & 31.30          & 59.54          & 71.80          & 21.15          & 29.18          & 58.66          & 70.23          & \textbf{17.02} \\
                              & Ours$^{\text{\textasteriskcentered}}$ & \textbf{33.26} & \textbf{66.70} & \textbf{76.92} & \textbf{18.77} & \textbf{32.20} & \textbf{60.14} & \textbf{71.96} & 17.05 \\
    \cdashline{1-10}
    \multirow{2}{*}{CC12M}    & CLIP$^{\text{\textasteriskcentered}}$ & 35.22          & 62.74          & 73.46          & 17.70          & 33.36          & 61.15          & 73.36          & 15.76 \\
                              & Ours$^{\text{\textasteriskcentered}}$ & \textbf{38.66} & \textbf{66.74} & \textbf{77.10} & \textbf{13.35} & \textbf{38.15} & \textbf{65.85} & \textbf{77.02} & \textbf{12.54} \\
    \bottomrule
    \end{tabular}
    \caption{\textbf{Image-Text Retrieval -- }Zero-shot retrieval performance on the COCO 5K testing set as measured by recall at 1,5,10 (higher is better) and mean rank (lower is better). The last four rows (marked with $^{*}$) report the zero-shot retrieval results of the same models further finetuned on the COCO captions training set.}
    \vspace{-2mm}
    \label{tab:retrieval}
	}
\end{table*}

Table \ref{tab:retrieval} shows our zero-shot retrieval performance relative to baseline CLIP under COCO, CC$3$M and CC$12$M pretraining datasets.
Relative to CLIP, our method consistently outperforms in both text-to-image and image-to-text subtasks, with both methods generally achieving slightly higher performance in text-to-image sub-task.
Note that, due to data domain differences, pretraining directly on COCO produces models with higher downstream performance than pretraining on CC$12$M despite the $100\times$ increase in data-size.
The last four rows show that, as expected, finetuning Conceptual Captions pretrained models on COCO produces retrieval performance that exceeds corresponding baselines and the performance from training on COCO alone.

\subsection{Ablation Study}
\noindent
We now investigate the contributions of individual components of our method, specifically the swapped prediction strategy for estimating soft-alignments over the forward bootstrapping strategy, dynamic minibatch partitioning, and progressive self-distillation to downstream performance. Following pretraining on COCO, we measure models' performance on COCO zero-shot image-text retrieval and ImageNet zero-shot classification.
\newcommand{\greencheck}{{\color{green}\checkmark}}
\newcommand{\redx}{{\textcolor{red}\xmark}}
\begin{table}[H]
	\small{
    \centering
    \begin{tabular}{ccccc}
    \toprule
    \multicolumn{3}{c}{\textbf{Method}} & \multicolumn{2}{c}{\textbf{Performance}} \\
    \cmidrule(lr){1-3} \cmidrule(lr){4-5}
    \shortstack[c]{Prediction\\ Mode} &
    \shortstack[c]{Dynamic \\ Partitioning} & 
    \shortstack[c]{Progressive\\ Distillation}   &
    \shortstack[c]{IN \\ Top-1} &
    \shortstack[c]{MS \\R1} \\
    \hline
    Baseline & -                 & -              & 8.18 & 27.58 \\
    \cdashline{1-5}
    Forward        & \redx       & \redx          & 6.23 & 20.40 \\
    Swapped        & \redx       & \redx          & 8.46 & 23.51 \\
    Forward        & \greencheck & \redx          & 8.37 & 25.64 \\
    Swapped        & \greencheck & \redx          & 8.81 & 26.24 \\
    Forward        & \greencheck & \greencheck    & 8.87 & 26.18 \\
    Swapped        & \greencheck & \greencheck    & 9.51 & 28.48 \\
    \bottomrule
    \end{tabular}
    \caption{\textbf{Abalation Study -- }Our results using COCO for pretraining. IN Top-1 is zero-shot ImageNet accuracy and MS R1 is mean recall at rank 1 across image-to-text and text-to-image subtasks.}
    \vspace{-0.2cm}
    \label{tab:ablations}
	}
\end{table}

Table \ref{tab:ablations} shows that directly using the model's posterior (bootstrapping) to assign alignments to a static subset of inputs leads to a degradation of learned representations and decreased performance compared to the baseline.
Here the teacher network is likely reinforcing its own mistakes without penalty from the loss, leading to partial representation collapse for the unaligned examples.
Utilizing the swapped prediction strategy restores zero-shot performance to slightly above baseline suggesting effective method for addressing this issue; however, retrieval performance is still lower than the baseline.
By incorporating a dynamic partitioning strategy, our method exceeds baseline performance.

\subsection{Qualitative Analysis}
\begin{figure}[t!]
    \centering
    \includegraphics[width=1.0\linewidth]{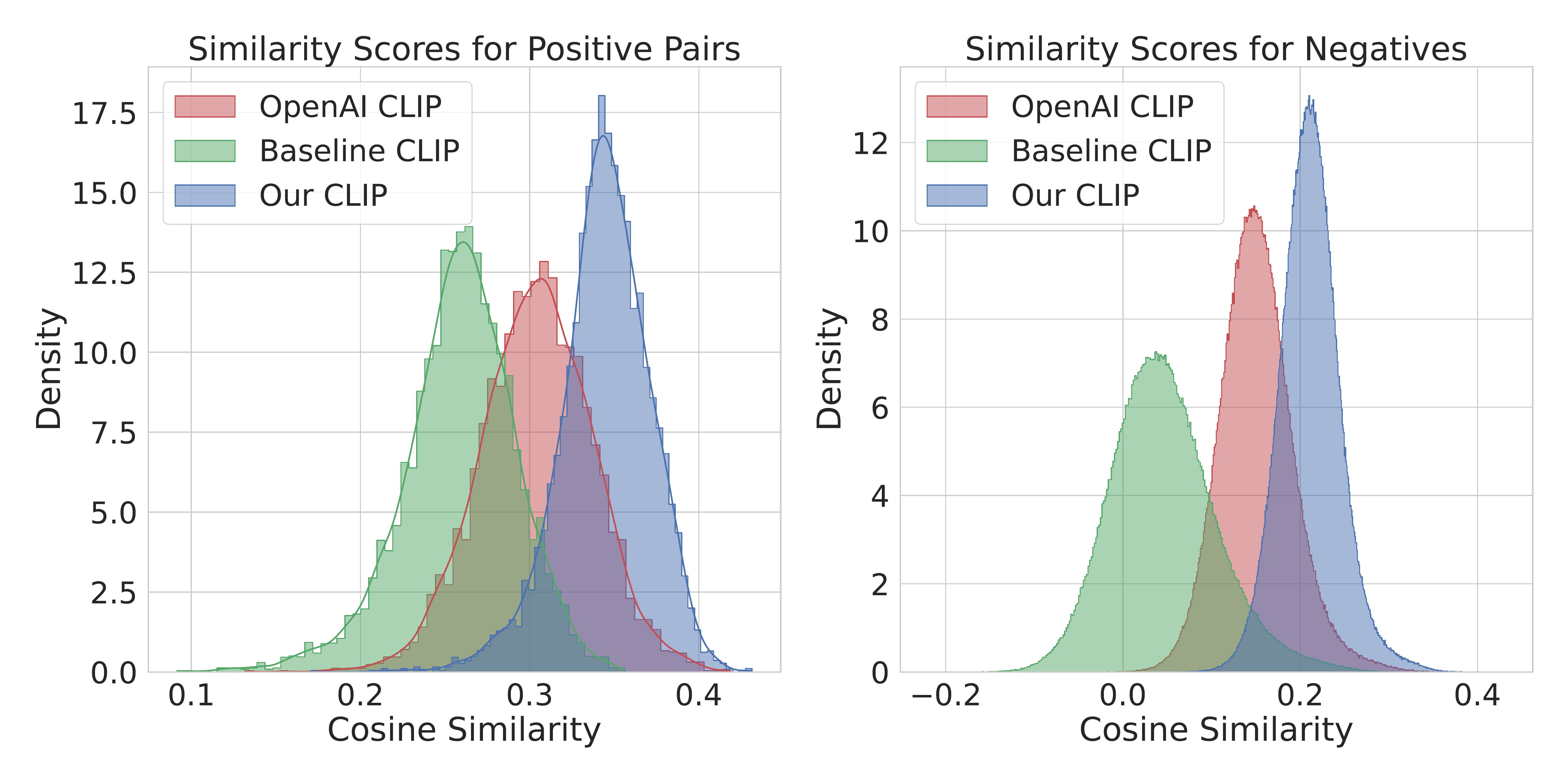}
    \caption{
    \small
    \textbf{Similarity Scores -- }Similarity scores distribution for positive and negative pairs in joint cross-modal space after training with original CLIP loss and our proposed loss is provided.
    Compared to our baseline (COCO) pretrained CLIP and OpenAI's pretrained CLIP model, our method yields similarity scores with higher mean and lower variance for positive pairs.
    While COCO pretrained CLIP model concentrates negatives' similarity scores around zero, our method concentrates them at higher levels as it allows for some degree of semantic similarity between negatives.
    }
    \vspace{-0.3cm}
    \label{fig:similarity_analysis}
\end{figure}

\myparagraph{a. Distribution of Similarity Scores:} Our method is largely motivated by the observation that previous VLP approaches neglect potential semantic similarity between negative samples and that accounting for this phenomenon can improve learned representations.
In Figure \ref{fig:similarity_analysis}, we plot the distribution of similarity scores for both positive and negative samples drawn from the COCO testing set.
The left subplot reveals that our method consistently yields larger similarity scores for positive pairs compared to its CLIP counterpart and the OpenAI pretrained CLIP.
Interestingly, the histogram of negative similarity scores shows that our method also assigns higher similarity scores to negative pairs.
While it may seem counter intuitive to assign greater similarity scores to negative samples, we argue that doing so is the very reason our method captures greater similarity between positive pairs.
By allowing some degree of alignment between the right set of negative examples, our method is able to minimize the inconsistencies between shared context of related positives and negatives.
This in turn allows us to learn an overall more coherent representation space, resulting in increased robustness and downstream performance.

\begin{figure}[ht!]
    \centering
    \includegraphics[width=0.99\linewidth]{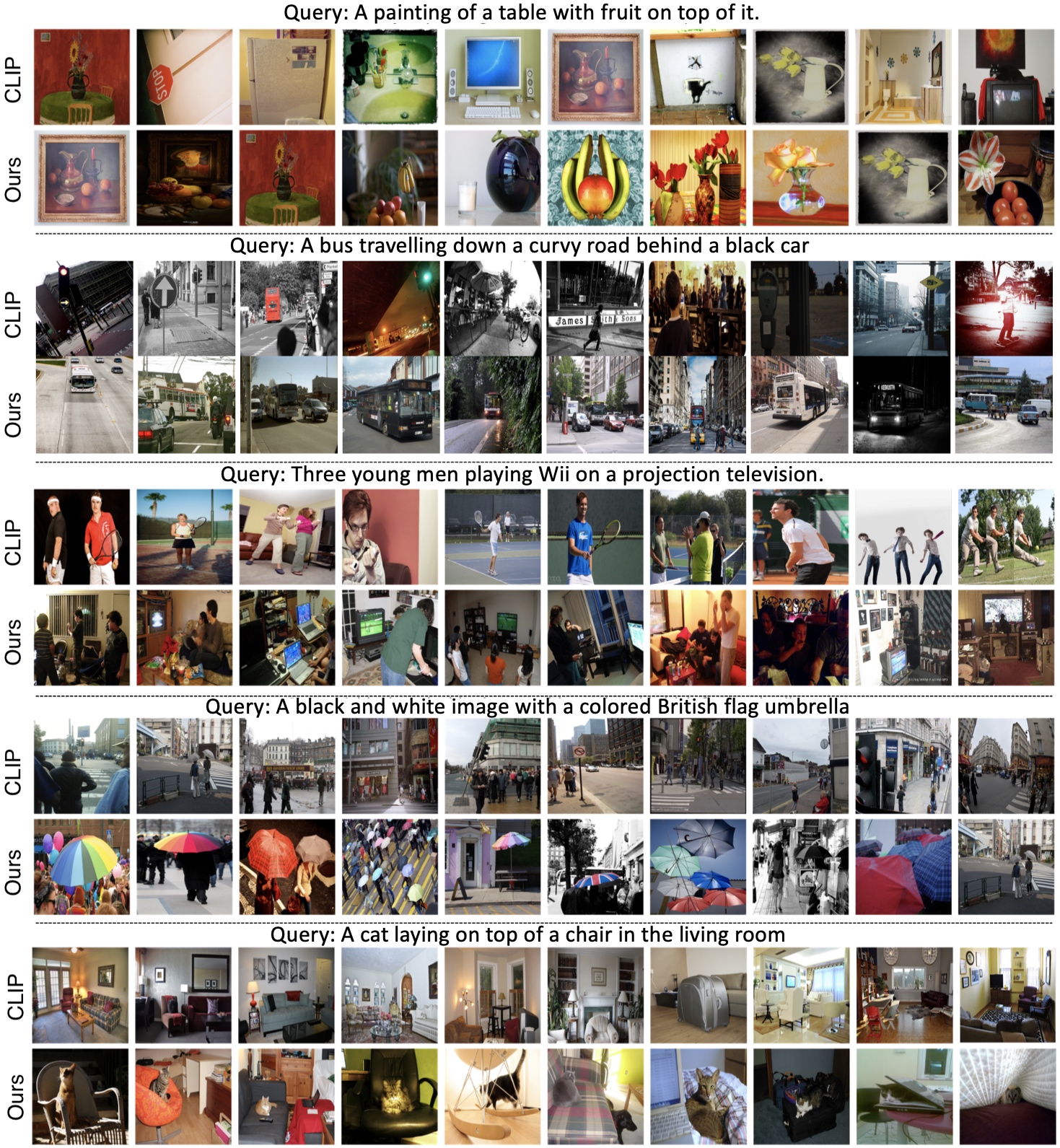}
    \caption{
    \small
    \textbf{Example Text-Image Retrievals --} Given a text query, we display the top ten most semantically related images (ranked left to right) retrieved by CLIP and our method. Compared to CLIP, our method continues to retrieve images that more holistically match the text description, even after the ground truth image has appeared in the ranking. 
    \vspace{-5mm}
    }
    \label{fig:qual}
\end{figure}

\myparagraph{b. Visualizing Text-Image Retrievals:} In Figure \ref{fig:qual}, we show the comparative list of top ten retrieved images for $5$ example text queries. Overall, these retrievals suggest that our learned features more comprehensively capture all potential similarities between a text snippet and image. In contrast, baseline CLIP features tend to more narrowly focus on one specific commonality while neglecting others. For example, in the last row of Figure \ref{fig:qual}, CLIP strongly targets ``living room'' but misses the cat. Our approach on the other hand is successfully able to extract all the key aspects of the query including the cat, the chair and the living room. Similar trends are found in other examples.

\vspace{-2mm}
\section{Conclusions}
\vspace{-2.0mm}
\noindent

\noindent We proposed a novel cross-modal contrastive learning framework with progressive self-distillation and soft image-text alignments.
Our approach distills its own knowledge to dynamically generate soft-alignment targets for a subset of samples in every minibatch, which enables it to efficiently learn robust representations from noisy data.
Extensive evaluations across $14$ benchmark datasets showed that our method consistently outperforms its CLIP counterpart in multiple settings.
Moreover, our method provides better effective robustness to natural distribution shifts compared to existing state-of-the-art methods.
Going forward, we plan to further improve the efficiency of our approach by investigating data redundancy, network architectures and optimization algorithms during pretraining, so that we can generalize to even larger scale with fewer resources.

\clearpage
{\small
\bibliographystyle{ieee_fullname}
\bibliography{main}

\begin{thebibliography}{10}\itemsep=-1pt

\bibitem{algan2021metalabelnet}
G{\"o}rkem Algan and Ilkay Ulusoy.
\newblock Metalabelnet: Learning to generate soft-labels from noisy-labels.
\newblock {\em arXiv preprint arXiv:2103.10869}, 2021.

\bibitem{arora2019theoretical}
Sanjeev Arora, Hrishikesh Khandeparkar, Mikhail Khodak, Orestis Plevrakis, and
  Nikunj Saunshi.
\newblock A theoretical analysis of contrastive unsupervised representation
  learning.
\newblock {\em arXiv preprint arXiv:1902.09229}, 2019.

\bibitem{barbu2019objectnet}
Andrei Barbu, David Mayo, Julian Alverio, William Luo, Christopher Wang, Dan
  Gutfreund, Josh Tenenbaum, and Boris Katz.
\newblock Objectnet: A large-scale bias-controlled dataset for pushing the
  limits of object recognition models.
\newblock In H. Wallach, H. Larochelle, A. Beygelzimer, F. d\textquotesingle
  Alch\'{e}-Buc, E. Fox, and R. Garnett, editors, {\em Advances in Neural
  Information Processing Systems}, volume~32. Curran Associates, Inc., 2019.

\bibitem{berg-birdsnap-cvpr2014}
Thomas Berg, Jiongxin Liu, Seung~Woo Lee, Michelle~L. Alexander, David~W.
  Jacobs, and Peter~N. Belhumeur.
\newblock Birdsnap: Large-scale fine-grained visual categorization of birds.
\newblock In {\em Proc. Conf. Computer Vision and Pattern Recognition (CVPR)},
  June 2014.

\bibitem{bossard14}
Lukas Bossard, Matthieu Guillaumin, and Luc Van~Gool.
\newblock Food-101 -- mining discriminative components with random forests.
\newblock In {\em European Conference on Computer Vision}, 2014.

\bibitem{bucilua2006model}
Cristian Buciluǎ, Rich Caruana, and Alexandru Niculescu-Mizil.
\newblock Model compression.
\newblock In {\em Proceedings of the 12th ACM SIGKDD international conference
  on Knowledge discovery and data mining}, pages 535--541, 2006.

\bibitem{caron2018deep}
Mathilde Caron, Piotr Bojanowski, Armand Joulin, and Matthijs Douze.
\newblock Deep clustering for unsupervised learning of visual features.
\newblock In {\em Proceedings of the European Conference on Computer Vision
  (ECCV)}, pages 132--149, 2018.

\bibitem{caron2020unsupervised}
Mathilde Caron, Ishan Misra, Julien Mairal, Priya Goyal, Piotr Bojanowski, and
  Armand Joulin.
\newblock Unsupervised learning of visual features by contrasting cluster
  assignments.
\newblock {\em arXiv preprint arXiv:2006.09882}, 2020.

\bibitem{changpinyo2021conceptual}
Soravit Changpinyo, Piyush Sharma, Nan Ding, and Radu Soricut.
\newblock Conceptual 12m: Pushing web-scale image-text pre-training to
  recognize long-tail visual concepts.
\newblock In {\em Proceedings of the IEEE/CVF Conference on Computer Vision and
  Pattern Recognition}, pages 3558--3568, 2021.

\bibitem{chen2017learning}
Guobin Chen, Wongun Choi, Xiang Yu, Tony Han, and Manmohan Chandraker.
\newblock Learning efficient object detection models with knowledge
  distillation.
\newblock {\em Advances in neural information processing systems}, 30, 2017.

\bibitem{chen2020simple}
Ting Chen, Simon Kornblith, Mohammad Norouzi, and Geoffrey Hinton.
\newblock A simple framework for contrastive learning of visual
  representations.
\newblock In {\em International Conference on Machine Learning (ICML)}, 2020.

\bibitem{deng2009imagenet}
Jia Deng, Wei Dong, Richard Socher, Li-Jia Li, Kai Li, and Li Fei-Fei.
\newblock Imagenet: A large-scale hierarchical image database.
\newblock In {\em IEEE Conference on Computer Vision and Pattern Recognition
  (CVPR)}. Ieee, 2009.

\bibitem{deshpande2015learning}
Aditya Deshpande, Jason Rock, and David Forsyth.
\newblock Learning large-scale automatic image colorization.
\newblock In {\em Proceedings of the IEEE International Conference on Computer
  Vision}, pages 567--575, 2015.

\bibitem{ding2019adaptive}
Qianggang Ding, Sifan Wu, Hao Sun, Jiadong Guo, and Shu-Tao Xia.
\newblock Adaptive regularization of labels.
\newblock {\em arXiv preprint arXiv:1908.05474}, 2019.

\bibitem{doersch2015unsupervised}
Carl Doersch, Abhinav Gupta, and Alexei~A Efros.
\newblock Unsupervised visual representation learning by context prediction.
\newblock In {\em Proceedings of the IEEE international conference on computer
  vision}, pages 1422--1430, 2015.

\bibitem{dosovitskiy2015discriminative}
Alexey Dosovitskiy, Philipp Fischer, Jost~Tobias Springenberg, Martin
  Riedmiller, and Thomas Brox.
\newblock Discriminative unsupervised feature learning with exemplar
  convolutional neural networks.
\newblock {\em IEEE transactions on pattern analysis and machine intelligence},
  38(9):1734--1747, 2015.

\bibitem{dosovitskiy2014discriminative}
Alexey Dosovitskiy, Jost~Tobias Springenberg, Martin Riedmiller, and Thomas
  Brox.
\newblock Discriminative unsupervised feature learning with convolutional
  neural networks.
\newblock {\em Advances in neural information processing systems}, 27:766--774,
  2014.

\bibitem{fei2006one}
Li Fei-Fei, Rob Fergus, and Pietro Perona.
\newblock One-shot learning of object categories.
\newblock {\em IEEE transactions on pattern analysis and machine intelligence},
  28(4):594--611, 2006.

\bibitem{furst2021cloob}
Andreas F{\"u}rst, Elisabeth Rumetshofer, Viet Tran, Hubert Ramsauer, Fei Tang,
  Johannes Lehner, David Kreil, Michael Kopp, G{\"u}nter Klambauer, Angela
  Bitto-Nemling, et~al.
\newblock Cloob: Modern hopfield networks with infoloob outperform clip.
\newblock {\em arXiv preprint arXiv:2110.11316}, 2021.

\bibitem{gidaris2018unsupervised}
Spyros Gidaris, Praveer Singh, and Nikos Komodakis.
\newblock Unsupervised representation learning by predicting image rotations.
\newblock {\em arXiv preprint arXiv:1803.07728}, 2018.

\bibitem{hahn2019self}
Sangchul Hahn and Heeyoul Choi.
\newblock Self-knowledge distillation in natural language processing.
\newblock {\em arXiv preprint arXiv:1908.01851}, 2019.

\bibitem{han2018co}
Bo Han, Quanming Yao, Xingrui Yu, Gang Niu, Miao Xu, Weihua Hu, Ivor Tsang, and
  Masashi Sugiyama.
\newblock Co-teaching: Robust training of deep neural networks with extremely
  noisy labels.
\newblock {\em arXiv preprint arXiv:1804.06872}, 2018.

\bibitem{han2019video}
Tengda Han, Weidi Xie, and Andrew Zisserman.
\newblock Video representation learning by dense predictive coding.
\newblock In {\em Proceedings of the IEEE/CVF International Conference on
  Computer Vision Workshops}, pages 0--0, 2019.

\bibitem{he2019moco}
Kaiming He, Haoqi Fan, Yuxin Wu, Saining Xie, and Ross Girshick.
\newblock Momentum contrast for unsupervised visual representation learning.
\newblock In {\em IEEE Conference on Computer Vision and Pattern Recognition
  (CVPR)}, 2019.

\bibitem{hendrycks2021many}
Dan Hendrycks, Steven Basart, Norman Mu, Saurav Kadavath, Frank Wang, Evan
  Dorundo, Rahul Desai, Tyler Zhu, Samyak Parajuli, Mike Guo, Dawn Song, Jacob
  Steinhardt, and Justin Gilmer.
\newblock The many faces of robustness: A critical analysis of
  out-of-distribution generalization.
\newblock {\em ICCV}, 2021.

\bibitem{hendrycks2019natural}
Dan Hendrycks, Kevin Zhao, Steven Basart, Jacob Steinhardt, and Dawn Song.
\newblock Natural adversarial examples.(2019).
\newblock {\em arXiv preprint cs.LG/1907.07174}, 2019.

\bibitem{hinton2015distilling}
Geoffrey Hinton, Oriol Vinyals, and Jeff Dean.
\newblock Distilling the knowledge in a neural network.
\newblock {\em arXiv preprint arXiv:1503.02531}, 2015.

\bibitem{ilharco_gabriel_2021_5143773}
Gabriel Ilharco, Mitchell Wortsman, Nicholas Carlini, Rohan Taori, Achal Dave,
  Vaishaal Shankar, Hongseok Namkoong, John Miller, Hannaneh Hajishirzi, Ali
  Farhadi, and Ludwig Schmidt.
\newblock Openclip, July 2021.
\newblock If you use this software, please cite it as below.

\bibitem{jason2016back}
J~Yu Jason, Adam~W Harley, and Konstantinos~G Derpanis.
\newblock Back to basics: Unsupervised learning of optical flow via brightness
  constancy and motion smoothness.
\newblock In {\em European Conference on Computer Vision}, pages 3--10.
  Springer, 2016.

\bibitem{jia2021scaling}
Chao Jia, Yinfei Yang, Ye Xia, Yi-Ting Chen, Zarana Parekh, Hieu Pham, Quoc~V
  Le, Yunhsuan Sung, Zhen Li, and Tom Duerig.
\newblock Scaling up visual and vision-language representation learning with
  noisy text supervision.
\newblock {\em arXiv preprint arXiv:2102.05918}, 2021.

\bibitem{jing2020self}
Longlong Jing and Yingli Tian.
\newblock Self-supervised visual feature learning with deep neural networks: A
  survey.
\newblock {\em IEEE transactions on pattern analysis and machine intelligence},
  2020.

\bibitem{kim2019self}
Dahun Kim, Donghyeon Cho, and In~So Kweon.
\newblock Self-supervised video representation learning with space-time cubic
  puzzles.
\newblock In {\em Proceedings of the AAAI Conference on Artificial
  Intelligence}, volume~33, pages 8545--8552, 2019.

\bibitem{kingma2014adam}
Diederik~P Kingma and Jimmy Ba.
\newblock Adam: A method for stochastic optimization.
\newblock {\em arXiv preprint arXiv:1412.6980}, 2014.

\bibitem{krizhevsky2009learning}
Alex Krizhevsky, Geoffrey Hinton, et~al.
\newblock Learning multiple layers of features from tiny images.
\newblock 2009.

\bibitem{larsson2016learning}
Gustav Larsson, Michael Maire, and Gregory Shakhnarovich.
\newblock Learning representations for automatic colorization.
\newblock In {\em European conference on computer vision}, pages 577--593.
  Springer, 2016.

\bibitem{lee2017unsupervised}
Hsin-Ying Lee, Jia-Bin Huang, Maneesh Singh, and Ming-Hsuan Yang.
\newblock Unsupervised representation learning by sorting sequences.
\newblock In {\em Proceedings of the IEEE International Conference on Computer
  Vision}, pages 667--676, 2017.

\bibitem{li2019learning}
Junnan Li, Yongkang Wong, Qi Zhao, and Mohan~S Kankanhalli.
\newblock Learning to learn from noisy labeled data.
\newblock In {\em Proceedings of the IEEE/CVF Conference on Computer Vision and
  Pattern Recognition}, pages 5051--5059, 2019.

\bibitem{li2020few}
Tianhong Li, Jianguo Li, Zhuang Liu, and Changshui Zhang.
\newblock Few sample knowledge distillation for efficient network compression.
\newblock In {\em Proceedings of the IEEE/CVF Conference on Computer Vision and
  Pattern Recognition}, pages 14639--14647, 2020.

\bibitem{li2021supervision}
Yangguang Li, Feng Liang, Lichen Zhao, Yufeng Cui, Wanli Ouyang, Jing Shao,
  Fengwei Yu, and Junjie Yan.
\newblock Supervision exists everywhere: A data efficient contrastive
  language-image pre-training paradigm.
\newblock {\em arXiv preprint arXiv:2110.05208}, 2021.

\bibitem{li2017learning}
Yuncheng Li, Jianchao Yang, Yale Song, Liangliang Cao, Jiebo Luo, and Li-Jia
  Li.
\newblock Learning from noisy labels with distillation.
\newblock In {\em Proceedings of the IEEE International Conference on Computer
  Vision}, pages 1910--1918, 2017.

\bibitem{lin2014microsoft}
Tsung-Yi Lin, Michael Maire, Serge Belongie, James Hays, Pietro Perona, Deva
  Ramanan, Piotr Doll{\'a}r, and C~Lawrence Zitnick.
\newblock Microsoft coco: Common objects in context.
\newblock In {\em European conference on computer vision}, pages 740--755.
  Springer, 2014.

\bibitem{lin2021self}
Yuanze Lin, Xun Guo, and Yan Lu.
\newblock Self-supervised video representation learning with meta-contrastive
  network.
\newblock In {\em Proceedings of the IEEE/CVF International Conference on
  Computer Vision}, pages 8239--8249, 2021.

\bibitem{liu2021inflate}
Haoliang Liu, Tan Yu, and Ping Li.
\newblock Inflate and shrink: Enriching and reducing interactions for fast
  text-image retrieval.
\newblock In {\em Proceedings of the 2021 Conference on Empirical Methods in
  Natural Language Processing}, pages 9796--9809, 2021.

\bibitem{loshchilov2016sgdr}
Ilya Loshchilov and Frank Hutter.
\newblock Sgdr: Stochastic gradient descent with warm restarts.
\newblock {\em arXiv preprint arXiv:1608.03983}, 2016.

\bibitem{micikevicius2017mixed}
Paulius Micikevicius, Sharan Narang, Jonah Alben, Gregory Diamos, Erich Elsen,
  David Garcia, Boris Ginsburg, Michael Houston, Oleksii Kuchaiev, Ganesh
  Venkatesh, et~al.
\newblock Mixed precision training.
\newblock {\em arXiv preprint arXiv:1710.03740}, 2017.

\bibitem{miller2021accuracy}
John~P Miller, Rohan Taori, Aditi Raghunathan, Shiori Sagawa, Pang~Wei Koh,
  Vaishaal Shankar, Percy Liang, Yair Carmon, and Ludwig Schmidt.
\newblock Accuracy on the line: On the strong correlation between
  out-of-distribution and in-distribution generalization.
\newblock In {\em International Conference on Machine Learning}, pages
  7721--7735. PMLR, 2021.

\bibitem{misra2020self}
Ishan Misra and Laurens van~der Maaten.
\newblock Self-supervised learning of pretext-invariant representations.
\newblock In {\em Proceedings of the IEEE/CVF Conference on Computer Vision and
  Pattern Recognition}, pages 6707--6717, 2020.

\bibitem{misra2016shuffle}
Ishan Misra, C~Lawrence Zitnick, and Martial Hebert.
\newblock Shuffle and learn: unsupervised learning using temporal order
  verification.
\newblock In {\em European Conference on Computer Vision}, pages 527--544.
  Springer, 2016.

\bibitem{mobahi2009deep}
Hossein Mobahi, Ronan Collobert, and Jason Weston.
\newblock Deep learning from temporal coherence in video.
\newblock In {\em Proceedings of the 26th Annual International Conference on
  Machine Learning}, pages 737--744, 2009.

\bibitem{morgado2021robust}
Pedro Morgado, Ishan Misra, and Nuno Vasconcelos.
\newblock Robust audio-visual instance discrimination.
\newblock In {\em Proceedings of the IEEE/CVF Conference on Computer Vision and
  Pattern Recognition}, pages 12934--12945, 2021.

\bibitem{morgado2020avid}
Pedro Morgado, Nuno Vasconcelos, and Ishan Misra.
\newblock Audio-visual instance discrimination with cross-modal agreement.
\newblock {\em CoRR}, abs/2004.12943, 2020.

\bibitem{morgado2021audio}
Pedro Morgado, Nuno Vasconcelos, and Ishan Misra.
\newblock Audio-visual instance discrimination with cross-modal agreement.
\newblock In {\em Proceedings of the IEEE/CVF Conference on Computer Vision and
  Pattern Recognition}, pages 12475--12486, 2021.

\bibitem{muller2019does}
Rafael M{\"u}ller, Simon Kornblith, and Geoffrey Hinton.
\newblock When does label smoothing help?
\newblock {\em arXiv preprint arXiv:1906.02629}, 2019.

\bibitem{noroozi2016unsupervised}
Mehdi Noroozi and Paolo Favaro.
\newblock Unsupervised learning of visual representations by solving jigsaw
  puzzles.
\newblock In {\em European conference on computer vision}, pages 69--84.
  Springer, 2016.

\bibitem{northcutt2021confident}
Curtis Northcutt, Lu Jiang, and Isaac Chuang.
\newblock Confident learning: Estimating uncertainty in dataset labels.
\newblock {\em Journal of Artificial Intelligence Research}, 70:1373--1411,
  2021.

\bibitem{oord2018representation}
Aaron van~den Oord, Yazhe Li, and Oriol Vinyals.
\newblock Representation learning with contrastive predictive coding.
\newblock {\em arXiv preprint arXiv:1807.03748}, 2018.

\bibitem{parkhi2012cats}
Omkar~M Parkhi, Andrea Vedaldi, Andrew Zisserman, and CV Jawahar.
\newblock Cats and dogs.
\newblock In {\em 2012 IEEE conference on computer vision and pattern
  recognition}, pages 3498--3505. IEEE, 2012.

\bibitem{pathak2017learning}
Deepak Pathak, Ross Girshick, Piotr Doll{\'a}r, Trevor Darrell, and Bharath
  Hariharan.
\newblock Learning features by watching objects move.
\newblock In {\em Proceedings of the IEEE Conference on Computer Vision and
  Pattern Recognition}, pages 2701--2710, 2017.

\bibitem{pathak2016context}
Deepak Pathak, Philipp Krahenbuhl, Jeff Donahue, Trevor Darrell, and Alexei~A
  Efros.
\newblock Context encoders: Feature learning by inpainting.
\newblock In {\em Proceedings of the IEEE conference on computer vision and
  pattern recognition}, pages 2536--2544, 2016.

\bibitem{patrini2017making}
Giorgio Patrini, Alessandro Rozza, Aditya Krishna~Menon, Richard Nock, and
  Lizhen Qu.
\newblock Making deep neural networks robust to label noise: A loss correction
  approach.
\newblock In {\em Proceedings of the IEEE conference on computer vision and
  pattern recognition}, pages 1944--1952, 2017.

\bibitem{pereyra2017regularizing}
Gabriel Pereyra, George Tucker, Jan Chorowski, {\L}ukasz Kaiser, and Geoffrey
  Hinton.
\newblock Regularizing neural networks by penalizing confident output
  distributions.
\newblock {\em arXiv preprint arXiv:1701.06548}, 2017.

\bibitem{piergiovanni2020evolving}
AJ Piergiovanni, Anelia Angelova, and Michael~S Ryoo.
\newblock Evolving losses for unsupervised video representation learning.
\newblock In {\em Proceedings of the IEEE/CVF Conference on Computer Vision and
  Pattern Recognition}, pages 133--142, 2020.

\bibitem{radford2021learning}
Alec Radford, Jong~Wook Kim, Chris Hallacy, Aditya Ramesh, Gabriel Goh,
  Sandhini Agarwal, Girish Sastry, Amanda Askell, Pamela Mishkin, Jack Clark,
  et~al.
\newblock Learning transferable visual models from natural language
  supervision.
\newblock {\em arXiv preprint arXiv:2103.00020}, 2021.

\bibitem{recht2019imagenet}
Benjamin Recht, Rebecca Roelofs, Ludwig Schmidt, and Vaishaal Shankar.
\newblock Do imagenet classifiers generalize to imagenet?
\newblock In {\em International Conference on Machine Learning}, pages
  5389--5400. PMLR, 2019.

\bibitem{reed2014training}
Scott Reed, Honglak Lee, Dragomir Anguelov, Christian Szegedy, Dumitru Erhan,
  and Andrew Rabinovich.
\newblock Training deep neural networks on noisy labels with bootstrapping.
\newblock {\em arXiv preprint arXiv:1412.6596}, 2014.

\bibitem{romero2014fitnets}
Adriana Romero, Nicolas Ballas, Samira~Ebrahimi Kahou, Antoine Chassang, Carlo
  Gatta, and Yoshua Bengio.
\newblock Fitnets: Hints for thin deep nets.
\newblock {\em arXiv preprint arXiv:1412.6550}, 2014.

\bibitem{sharma2018conceptual}
Piyush Sharma, Nan Ding, Sebastian Goodman, and Radu Soricut.
\newblock Conceptual captions: A cleaned, hypernymed, image alt-text dataset
  for automatic image captioning.
\newblock In {\em Proceedings of the 56th Annual Meeting of the Association for
  Computational Linguistics (Volume 1: Long Papers)}, pages 2556--2565, 2018.

\bibitem{shen2021much}
Sheng Shen, Liunian~Harold Li, Hao Tan, Mohit Bansal, Anna Rohrbach, Kai-Wei
  Chang, Zhewei Yao, and Kurt Keutzer.
\newblock How much can clip benefit vision-and-language tasks?
\newblock {\em arXiv preprint arXiv:2107.06383}, 2021.

\bibitem{shu2019meta}
Jun Shu, Qi Xie, Lixuan Yi, Qian Zhao, Sanping Zhou, Zongben Xu, and Deyu Meng.
\newblock Meta-weight-net: Learning an explicit mapping for sample weighting.
\newblock {\em arXiv preprint arXiv:1902.07379}, 2019.

\bibitem{taori2020measuring}
Rohan Taori, Achal Dave, Vaishaal Shankar, Nicholas Carlini, Benjamin Recht,
  and Ludwig Schmidt.
\newblock Measuring robustness to natural distribution shifts in image
  classification.
\newblock In {\em Advances in Neural Information Processing Systems (NeurIPS)},
  2020.

\bibitem{thomee2016yfcc100m}
Bart Thomee, David~A Shamma, Gerald Friedland, Benjamin Elizalde, Karl Ni,
  Douglas Poland, Damian Borth, and Li-Jia Li.
\newblock Yfcc100m: The new data in multimedia research.
\newblock {\em Communications of the ACM}, 59(2):64--73, 2016.

\bibitem{tian2020cmc}
Yonglong Tian, Dilip Krishnan, and Phillip Isola.
\newblock Contrastive multiview coding.
\newblock In {\em European Conference on Computer Vision (ECCV)}, 2020.

\bibitem{tian2020contrastive}
Yonglong Tian, Dilip Krishnan, and Phillip Isola.
\newblock Contrastive multiview coding.
\newblock In {\em Computer Vision--ECCV 2020: 16th European Conference,
  Glasgow, UK, August 23--28, 2020, Proceedings, Part XI 16}, pages 776--794.
  Springer, 2020.

\bibitem{vincent2008extracting}
Pascal Vincent, Hugo Larochelle, Yoshua Bengio, and Pierre-Antoine Manzagol.
\newblock Extracting and composing robust features with denoising autoencoders.
\newblock In {\em Proceedings of the 25th international conference on Machine
  learning}, pages 1096--1103, 2008.

\bibitem{wang2021efficientclip}
Jue Wang, Haofan Wang, Jincan Deng, Weijia Wu, and Debing Zhang.
\newblock Efficientclip: Efficient cross-modal pre-training by ensemble
  confident learning and language modeling.
\newblock {\em arXiv preprint arXiv:2109.04699}, 2021.

\bibitem{wang2019symmetric}
Yisen Wang, Xingjun Ma, Zaiyi Chen, Yuan Luo, Jinfeng Yi, and James Bailey.
\newblock Symmetric cross entropy for robust learning with noisy labels.
\newblock In {\em Proceedings of the IEEE/CVF International Conference on
  Computer Vision}, pages 322--330, 2019.

\bibitem{wu2018unsupervised}
Zhirong Wu, Yuanjun Xiong, X~Yu Stella, and Dahua Lin.
\newblock Unsupervised feature learning via non-parametric instance
  discrimination.
\newblock In {\em IEEE Conference on Computer Vision and Pattern Recognition
  (CVPR)}, 2018.

\bibitem{xu2019data}
Ting-Bing Xu and Cheng-Lin Liu.
\newblock Data-distortion guided self-distillation for deep neural networks.
\newblock In {\em Proceedings of the AAAI Conference on Artificial
  Intelligence}, volume~33, pages 5565--5572, 2019.

\bibitem{yin2018geonet}
Zhichao Yin and Jianping Shi.
\newblock Geonet: Unsupervised learning of dense depth, optical flow and camera
  pose.
\newblock In {\em Proceedings of the IEEE conference on computer vision and
  pattern recognition}, pages 1983--1992, 2018.

\bibitem{yun2020regularizing}
Sukmin Yun, Jongjin Park, Kimin Lee, and Jinwoo Shin.
\newblock Regularizing class-wise predictions via self-knowledge distillation.
\newblock In {\em Proceedings of the IEEE/CVF conference on computer vision and
  pattern recognition}, pages 13876--13885, 2020.

\bibitem{zhang2016colorful}
Richard Zhang, Phillip Isola, and Alexei~A Efros.
\newblock Colorful image colorization.
\newblock In {\em European conference on computer vision}, pages 649--666.
  Springer, 2016.

\bibitem{zhang2017split}
Richard Zhang, Phillip Isola, and Alexei~A Efros.
\newblock Split-brain autoencoders: Unsupervised learning by cross-channel
  prediction.
\newblock In {\em Proceedings of the IEEE Conference on Computer Vision and
  Pattern Recognition}, pages 1058--1067, 2017.

\bibitem{zhang2018generalized}
Zhilu Zhang and Mert~R Sabuncu.
\newblock Generalized cross entropy loss for training deep neural networks with
  noisy labels.
\newblock In {\em 32nd Conference on Neural Information Processing Systems
  (NeurIPS)}, 2018.

\bibitem{zhang2020distilling}
Zizhao Zhang, Han Zhang, Sercan~O Arik, Honglak Lee, and Tomas Pfister.
\newblock Distilling effective supervision from severe label noise.
\newblock In {\em Proceedings of the IEEE/CVF Conference on Computer Vision and
  Pattern Recognition}, pages 9294--9303, 2020.

\bibitem{zhou2017places}
Bolei Zhou, Agata Lapedriza, Aditya Khosla, Aude Oliva, and Antonio Torralba.
\newblock Places: A 10 million image database for scene recognition.
\newblock {\em IEEE Transactions on Pattern Analysis and Machine Intelligence},
  2017.

\bibitem{zhuang2020unsupervised}
Chengxu Zhuang, Tianwei She, Alex Andonian, Max~Sobol Mark, and Daniel Yamins.
\newblock Unsupervised learning from video with deep neural embeddings.
\newblock In {\em Proceedings of the IEEE/CVF Conference on Computer Vision and
  Pattern Recognition}, pages 9563--9572, 2020.

\bibitem{zhuang2019local}
Chengxu Zhuang, Alex~Lin Zhai, and Daniel Yamins.
\newblock Local aggregation for unsupervised learning of visual embeddings.
\newblock In {\em Proceedings of the IEEE/CVF International Conference on
  Computer Vision}, pages 6002--6012, 2019.

\bibitem{zolfaghari2021crossclr}
Mohammadreza Zolfaghari, Yi Zhu, Peter Gehler, and Thomas Brox.
\newblock Crossclr: Cross-modal contrastive learning for multi-modal video
  representations.
\newblock In {\em Proceedings of the IEEE/CVF International Conference on
  Computer Vision}, pages 1450--1459, 2021.

\end{thebibliography}


\begin{thebibliography}{1}\itemsep=-1pt

\bibitem{bommasani2021opportunities}
Rishi Bommasani, Drew~A Hudson, Ehsan Adeli, Russ Altman, Simran Arora, Sydney
  von Arx, Michael~S Bernstein, Jeannette Bohg, Antoine Bosselut, Emma
  Brunskill, et~al.
\newblock On the opportunities and risks of foundation models.
\newblock {\em arXiv preprint arXiv:2108.07258}, 2021.

\bibitem{davenport2019potential}
Thomas Davenport and Ravi Kalakota.
\newblock The potential for artificial intelligence in healthcare.
\newblock {\em Future healthcare journal}, 6(2):94, 2019.

\bibitem{furst2021cloob}
Andreas F{\"u}rst, Elisabeth Rumetshofer, Viet Tran, Hubert Ramsauer, Fei Tang,
  Johannes Lehner, David Kreil, Michael Kopp, G{\"u}nter Klambauer, Angela
  Bitto-Nemling, et~al.
\newblock Cloob: Modern hopfield networks with infoloob outperform clip.
\newblock {\em arXiv preprint arXiv:2110.11316}, 2021.

\bibitem{ilharco_gabriel_2021_5143773}
Gabriel Ilharco, Mitchell Wortsman, Nicholas Carlini, Rohan Taori, Achal Dave,
  Vaishaal Shankar, Hongseok Namkoong, John Miller, Hannaneh Hajishirzi, Ali
  Farhadi, and Ludwig Schmidt.
\newblock Openclip, July 2021.
\newblock If you use this software, please cite it as below.

\bibitem{ingle2016tesla}
Shantanu Ingle and Madhuri Phute.
\newblock Tesla autopilot: semi autonomous driving, an uptick for future
  autonomy.
\newblock {\em International Research Journal of Engineering and Technology},
  3(9):369--372, 2016.

\bibitem{li2021supervision}
Yangguang Li, Feng Liang, Lichen Zhao, Yufeng Cui, Wanli Ouyang, Jing Shao,
  Fengwei Yu, and Junjie Yan.
\newblock Supervision exists everywhere: A data efficient contrastive
  language-image pre-training paradigm.
\newblock {\em arXiv preprint arXiv:2110.05208}, 2021.

\bibitem{scikit-learn}
F. Pedregosa, G. Varoquaux, A. Gramfort, V. Michel, B. Thirion, O. Grisel, M.
  Blondel, P. Prettenhofer, R. Weiss, V. Dubourg, J. Vanderplas, A. Passos, D.
  Cournapeau, M. Brucher, M. Perrot, and E. Duchesnay.
\newblock Scikit-learn: Machine learning in {P}ython.
\newblock {\em Journal of Machine Learning Research}, 12:2825--2830, 2011.

\bibitem{radford2021learning}
Alec Radford, Jong~Wook Kim, Chris Hallacy, Aditya Ramesh, Gabriel Goh,
  Sandhini Agarwal, Girish Sastry, Amanda Askell, Pamela Mishkin, Jack Clark,
  et~al.
\newblock Learning transferable visual models from natural language
  supervision.
\newblock {\em arXiv preprint arXiv:2103.00020}, 2021.

\bibitem{yao2021filip}
Lewei Yao, Runhui Huang, Lu Hou, Guansong Lu, Minzhe Niu, Hang Xu, Xiaodan
  Liang, Zhenguo Li, Xin Jiang, and Chunjing Xu.
\newblock Filip: Fine-grained interactive language-image pre-training.
\newblock {\em arXiv preprint arXiv:2111.07783}, 2021.

\end{thebibliography}
}
\end{document}


\title{Robust Cross-Modal Representation Learning with Progressive Self-Distillation\\ Supplementary Materials}

\author{Alex Andonian\thanks{This work was done when the author was an intern at Amazon.}\\
MIT CSAIL\\
{\tt\small andonian@mit.edu}
\and
Shixing Chen, Raffay Hamid\\
Amazon Prime Video\\
{\tt\small \{shixic, raffay\}@amazon.com}
}

\maketitle

\section{Summary of Datasets}

\subsection{Pretraining Datasets}

\noindent
Table \ref{tab:pretraining_datasets} lists the exact number of examples used for pretraining from each dataset. During COCO pretraining, we randomly select one of five unique captions assigned to each image, effectively multiplying the total number of image-text pairs by a factor of 5.
Unlike COCO, the Conceptual Captions datasets are not stored in one central location and not all of the provided URLs are still valid. Therefore, the actual number of pretraining examples is less than the advertised amount by as much as 1M+ in the case of CC12M.

\begin{table}[h!]
    \centering
    \begin{tabular}{c|ccc}
    \hline
    \textbf{Dataset} & \textbf{COCO} & \textbf{CC3M} & \textbf{CC12M} \\
    \hline
    \# & 118,287  & 2,884,940  & 10,707,814 \\
    \hline
    \end{tabular}
    \caption{\textbf{Pretraining Dataset Sizes --} Exact sizes of pretraining datasets employed in this work (no. of image-text pairs). }
    \label{tab:pretraining_datasets}
\end{table}

\subsection{Evaluation Datasets}
\noindent
In Table~\ref{tab:eval_datasets}, we list additional details for the evaluation datasets in this study including the number of classes and the sizes of the training-testing splits. For the last 5 rows (ObjectNet and ImageNet-\{A,O,R,V2\}), we list only the number of classes and the testing set size as these have been designated as ``testing-only'' datasets.
\begin{table}[h!]
    \centering
    \begin{tabular}{lccc}
    \hline
    \textbf{Dataset} & \textbf{Classes} & \textbf{Train Size} & \textbf{Test Size} \\
    \hline
    Cifar10    & 10   & 50,000    & 10,000 \\
    Cifar100   & 100  & 50,000    & 10,000 \\
    Caltech101 & 102  & 2,863     & 8,677 \\
    Food101    & 101  & 75,750    & 25,250 \\
    OxfordPets & 37   & 3,680     & 3,669  \\
    Birdsnap   & 500  & 38,344    & 1,900  \\
    ImageNet   & 1000 & 1,281,167 & 50,000 \\
    Places365  & 365  & 1,803,460 & 36,489 \\
    ObjectNet  & 313  & -         & 50,000 \\
    ImageNet-A & 200  & -         & 7,500  \\
    ImageNet-O & 200  & -         & 2,000  \\
    ImageNet-R & 200  & -         & 30,000 \\
    ImageNetV2 & 1000 & -         & 30,000 \\
    
    \hline
    \end{tabular}
    \caption{\textbf{Evaluation Dataset Details --} The number of classes, training and testing examples present in the evaluation datasets. The size of the training set size for the last 5 rows is omitted because these datasets are designated as ``testing-only'' benchmarks.}
    \label{tab:eval_datasets}
\end{table}

\section{Detailed Experimental Settings}
\noindent

\subsection{Implementation Details}
\noindent
Our pretraining implementation largely follows CLIP~\cite{radford2021learning} with significant deviations motivated by computational constraints or empirical observations.
Table~\ref{tab:hparams} summarizes common hyperparameters settings shared across experiments.
Notable differences from \cite{radford2021learning} include a reduced batch size, learning rate and weight decay, but increased number of training epochs and warm-up iterations.
Unlike CLIP, which computes sharded, intra-GPU embedding similarities only, we perform a global all-gather operation to compute all pair-wise similarities within a batch.

\begin{table}[h!]
    \centering
    \begin{tabular}{lc}
    \toprule
    \textbf{Hyperparameter}            & \textbf{Value}     \\
    \midrule
    Batch size                         & 4096               \\
    Vocabulary size                    & 49408              \\
    Training epochs                    & 100                \\
    Initial temperature $\tau$         & 0.07               \\
    Teacher temperature $\tilde{\tau}$ & 0.1                \\
    Weight decay                       & 0.001              \\
    Warm-up iterations (\%)            & 0.2                \\
    Learning rate                      & $1 \times 10^{-5}$ \\
    Adam $\beta_1$                     & 0.9                \\
    Adam $\beta_2$                     & 0.99               \\
    Adam $\epsilon$                    & $10^{-5}$          \\
    \bottomrule
    \end{tabular}
    \caption{\textbf{Common hyperparameters --} used for both baseline CLIP pretraining and our method.}
    \label{tab:hparams}
	
\end{table}

\subsection{Complexity Analysis}
\noindent
In order to underscore how our method improves data-efficiency without incurring additional computation cost, we note that the number of trainable parameters in the ViT-B/32 (151.3M) and RN50 (102.0M) CLIP models are precisely the same under our method as the architectures remain unmodified with no additional parameters required. Also, the average time per training iteration (\emph{e.g.,} 0.253s for our COCO experiments using 8 Nvidia V100 GPUs) and memory requirements are effectively identical as well. 
Finally, we contrast our self-distillation approach to traditional knowledge-distillation methods and other choices of teacher network (\emph{e.g.,} a momentum teacher design), which may provide similar data-efficiency improvements, but incur significant computational costs of at least 2x compute and memory requirements or more.

\subsection{Prompt Engineering}
\noindent
Table~\ref{tab:prompts} lists the prompt templates used for zero-shot classification on the evaluation datasets. For each template, string interpolation replaces the placeholder symbol \{\} with a text representation of the category name, and a grammatical correction is applied to the preceding article, i.e., \texttt{a} $\rightarrow$ \texttt{an} for categories that start with a vowel.

\begin{table*}[h!]
    \centering
    \begin{tabular}[t!]{l|l|l}
    \hline
    \small
    \textbf{Cifar\{10,100\}} &
    \textbf{Caltech101} &
    \textbf{ImageNet+}              \\
    \hline
    \scriptsize{\shortstack[l]{
    ''a photo of a \{\}.'',\\
    ''a blurry photo of a \{\}.'',\\
    ''a black and white photo of a \{\}.'',\\
    ''a low contrast photo of a \{\}.'',\\
    ''a high contrast photo of a \{\}.'',
    ''a bad photo of a \{\}.'',\\
    ''a good photo of a \{\}.'',
    ''a photo of a small \{\}.'',\\
    ''a photo of a big \{\}.'',
    ''a photo of the \{\}.'',\\
    ''a blurry photo of the \{\}.'',
    ''a black and white photo of the \{\}.'',\\
    ''a low contrast photo of the \{\}.'',
    ''a high contrast photo of the \{\}.'',\\
    ''a bad photo of the \{\}.'',
    ''a good photo of the \{\}.'',\\
    ''a photo of the small \{\}.'',
    ''a photo of the big \{\}.''
    }}
    &
    \scriptsize{
    \shortstack[l]{
    ``a photo of a \{\}.'',
    ``a painting of a \{\}.'',
    ``a plastic \{\}.'',\\
    ``a sculpture of a \{\}.'',
    ``a sketch of a \{\}.'', 
    ``a tattoo of a \{\}.'',\\
    ``a toy \{\}.'',
    ``a rendition of a \{\}.'',
    ``a embroidered \{\}.'',
    ``a cartoon \{\}.'',\\
    ``a \{\} in a video game.'',
    ``a plushie \{\}.'',
    ``a origami \{\}.'',
    ``art of a \{\}.'',\\
    ``graffiti of a \{\}.'',
    ``a drawing of a \{\}.'',
    ``a doodle of a \{\}.'',\\
    ``a photo of the \{\}.'',
    ``a painting of the \{\}.'',
    ``the plastic \{\}.'',\\
    ``a sculpture of the \{\}.'',
    ``a sketch of the \{\}.'',
    ``a tattoo of the \{\}.'',\\
    ``the toy \{\}.'',
    ``a rendition of the \{\}.'',
    ``the embroidered \{\}.'',\\
    ``the cartoon \{\}.'',
    ``the \{\} in a video game.'',
    ``the plushie \{\}.'',\\
    ``the origami \{\}.'',
    ``art of the \{\}.'',
    ``graffiti of the \{\}.'',\\
    ``a drawing of the \{\}.'',
    ``a doodle of the \{\}.''
    }}
    &
    \footnotesize{
    \shortstack[l]{
    ``\{\}''\\
    ``A photo of \{\}''\\
    ``A photo the \{\}''\\
    ``itap of a \{\}.''\\
    ``a bad photo of the \{\}.''\\
    ``a origami \{\}.''\\
    ``a photo of the large \{\}.''\\
    ``a \{\} in a video game.''\\
    ``art of the \{\}.''\\
    ``a photo of the small \{\}.''
	}} \\
    \hline
    \end{tabular}
    \caption{\textbf{Prompt templates for zero-shot evaluation --} The placeholder symbol \{\} is replaced with a string representation of the category name. The last column ``ImageNet+'' corresponds to the templates used for all other datasets that appear in this work, including all of the ImageNet variants. }
    \label{tab:prompts}
	
\end{table*}

\subsection{Linear Probe Details}
\noindent
The linear probe evaluation involves training a logistic regression classifier on the frozen visual features extracted using the model's image encoder. Following CLIP~\cite{radford2021learning}, we train the logistic regression for a maximum of 1000 iterations using the L-BFGS optimization algorithm provided by scikit-learn~\cite{scikit-learn}.
We use the train/test split sizes listed in Table~\ref{tab:eval_datasets}.

\section{Comparison to Concurrent Works}

\noindent In this section, we provide a comparison of our work to unpublished concurrent works.

\noindent
While the concurrent works discussed in this section are aimed at improving on CLIP, they altogether vary along several dimensions including pretraining dataset, backbone architecture, hyperparameters and training details.
Due to the large number of unique experimental configurations, it is not feasible to precisely replicate the setting of each concurrent work; therefore, we provide our most closely matched experiments while acknowledging that the comparisons are not exactly one-to-one analogs.

Table~\ref{tab:zeroshot_comparison} lists the commonly reported ImageNet zero-shot Top 1 accuracy achieved by concurrent methods aimed at reproducing CLIP and/or addressing its limitations.
Each work additionally provides a re-implementation of CLIP (listed as CLIP impl. followed by a citation).
Even for architecture and dataset matched experiments, the difference in accuracy can differ by as much as 4.8\% (compare \cite{li2021supervision} and \cite{radford2021learning}), further highlighting the challenges of providing meaningful comparison.
From the perspective of raw performance, our method achieves the highest absolute top 1 accuracy on zero-shot ImageNet classification out of all approaches and architectures that use at most 15M pretraining examples.

\begin{table}[H]
	\small{
    \centering
    \begin{tabular}{lllc}
    \toprule
    \shortstack[c]{\textbf{Dataset}} &
    \shortstack[c]{\textbf{Method}} &
   \textbf{Architecture} &
   \textbf{ImageNet} \\
    \midrule
    \multirow{10}{*}{YFCC (15M)}     & CLIP impl.~\cite{yao2021filip}               & ViT-B/32 & 30.4 \\
                                     & FILIP~\cite{yao2021filip}                    & ViT-B/32 & 37.8 \\
                                     & CLOOB~\cite{furst2021cloob}                  & RN50     & 35.7 \\
                                     & CLOOB~\cite{furst2021cloob}                  & RN101    & 37.1 \\
                                     & CLOOB~\cite{furst2021cloob}                  & RN50x4   & 39.0 \\
                                     & CLIP impl.~\cite{li2021supervision}          & RN50     & 35.9 \\
                                     & DeCLIP~\cite{li2021supervision}              & RN50     & 41.9 \\
                                     & OpenAI CLIP~\cite{radford2021learning}       & RN50     & 31.3 \\
                                     & OpenCLIP~\cite{ilharco_gabriel_2021_5143773} & RN50     & 32.7 \\
                                     & OpenCLIP~\cite{ilharco_gabriel_2021_5143773} & RN101    & 34.8 \\
    \cdashline{1-4}
    \multirow{7}{*}{CC3M}     & OpenCLIP~\cite{ilharco_gabriel_2021_5143773} & RN50x4   & 22.2 \\
                              & CLIP impl.~\cite{furst2021cloob}             & RN50     & 23.9 \\ 
                              & CLOOB~\cite{furst2021cloob}                  & RN50     & 25.6 \\
                              & DeCLIP~\cite{li2021supervision}              & RN50     & 27.8 \\
                              & CLIP impl. (ours)                            & ViT-B/32 & 23.5 \\
                              & \textbf{Our method}                          & ViT-B/32 & \textbf{28.0} \\
    \cdashline{1-4}
    \multirow{3}{*}{CC12M}     & CLIP impl. (ours)               & ViT-B/32 & 37.8          \\
                               & DeCLIP~\cite{li2021supervision} & RN50     & 41.0          \\ 
                               & \textbf{Our method}             & ViT-B/32 & \textbf{42.2} \\
    \bottomrule
    \end{tabular}
    \caption{\textbf{Zero-Shot ImageNet Classification Comparison -- } Zero-shot Top1 accuracy (\%) of our method compared to concurrent works.}
    \label{tab:zeroshot_comparison}
	}
\end{table}

\section{Additional Quantitative Results}

\subsection{ResNet50 Visual Backbone Results}
\noindent
In order to demonstrate the effectiveness of our method to a broader experimental setting, we present results from experiments that utilize the widely adopted ResNet50 (RN50) visual backbone in Table \ref{tab:resultsRN50}. These results are highly consistent with performance trends observed with the ViT backbone (even with minimal hyperparameter tuning), suggesting that our method is amenable to CNN-based backbones as well.

\begin{table*}[!h]
    \vspace{-5mm}
	\small{
    \centering
    \begin{tabular}{ccccccccccccl}
    \small{\shortstack[c]{Pretraining\\ Dataset}} &
    \small{\shortstack[c]{Method}} &
    \rotatebox{70}{\tiny{Cifar10}} &
    \rotatebox{70}{\tiny{Cifar100}} &
    \rotatebox{70}{\tiny{Caltech101}} &
    \rotatebox{70}{\tiny{Places365}} &
    \rotatebox{70}{\tiny{ObjectNet}} &
    \rotatebox{70}{\tiny{ImageNet-R}} &
    \rotatebox{70}{\tiny{ImageNet-O}} &
    \rotatebox{70}{\tiny{Imagenet-A}} &
    \rotatebox{70}{\tiny{ImageNetV2}} &
    \rotatebox{70}{\tiny{ImageNet}} &
    \rotatebox{70}{\tiny{Average}} \\
    \midrule
    \multirow{2}{*}{COCO}     & CLIP & 31.12 & 7.66 & 27.01 & 10.56 & 2.64 & 5.77  & 8.9  & 2.31 & 5.92 & 6.59 & 10.84 \\
                              & Ours & \textbf{37.85} & \textbf{9.97} & \textbf{33.87} & \textbf{11.17} & \textbf{3.38}& \textbf{6.48} & \textbf{11.7}& \textbf{2.56}& \textbf{6.87}& \textbf{7.48}& $\textbf{13.12}^{\textbf{\textcolor{mygreen}{+2.27}}}$ \\
    \cdashline{1-13}
    \multirow{2}{*}{CC3M}     & CLIP & 36.46 & 14.20 & 44.81 & 19.29 & 3.24 & 19.28 & 24.50 & 3.53 & 16.95 & 18.12  & 20.04 \\
                              & Ours & \textbf{56.69} & \textbf{24.44} & \textbf{61.93} & \textbf{24.67} & \textbf{5.27} &\textbf{25.96} & \textbf{31.65} & \textbf{4.57} & \textbf{21.65} & \textbf{22.918} & $\textbf{27.98}^{\textbf{\textcolor{mygreen}{+7.94}}}$ \\
    \bottomrule
    \end{tabular}
    \caption{\textbf{Zero-Shot Image Classification Comparison with RN50 backbone -- } Zero-shot Top1 accuracy (\%) of our method compared to baseline CLIP on numerous downstream benchmark datasets. Note: Results are derived from published hyperparameters for the baseline and minimal hyperparameter tuning of our method to account for differences in backbone architecture.}
    \label{tab:resultsRN50}
	}
\end{table*}

\subsection{Ablation Studies for $\alpha$-scheduling}
\noindent
Our method is fairly robust to the choice of $\alpha$ scheduling. For example, replacing cosine annealing with a simple linear schedule produces a network with near identical performance on downstream evaluations. Specifically, for COCO the two scheduling methods result in an absolute mean difference in top1 classification of $0.13\%$ and relative difference of $0.67\%$ for our method using a ViT backbone.
Similarly, for CC3M we observe an absolute mean difference of $-0.22\%$ and relative difference of $-0.63\%$.

\section{Additional Qualitative Results}

\subsection{Additional Text-Image Retrievals}
\begin{figure*}[ht!]
    \centering
    \includegraphics[width=0.99\linewidth]{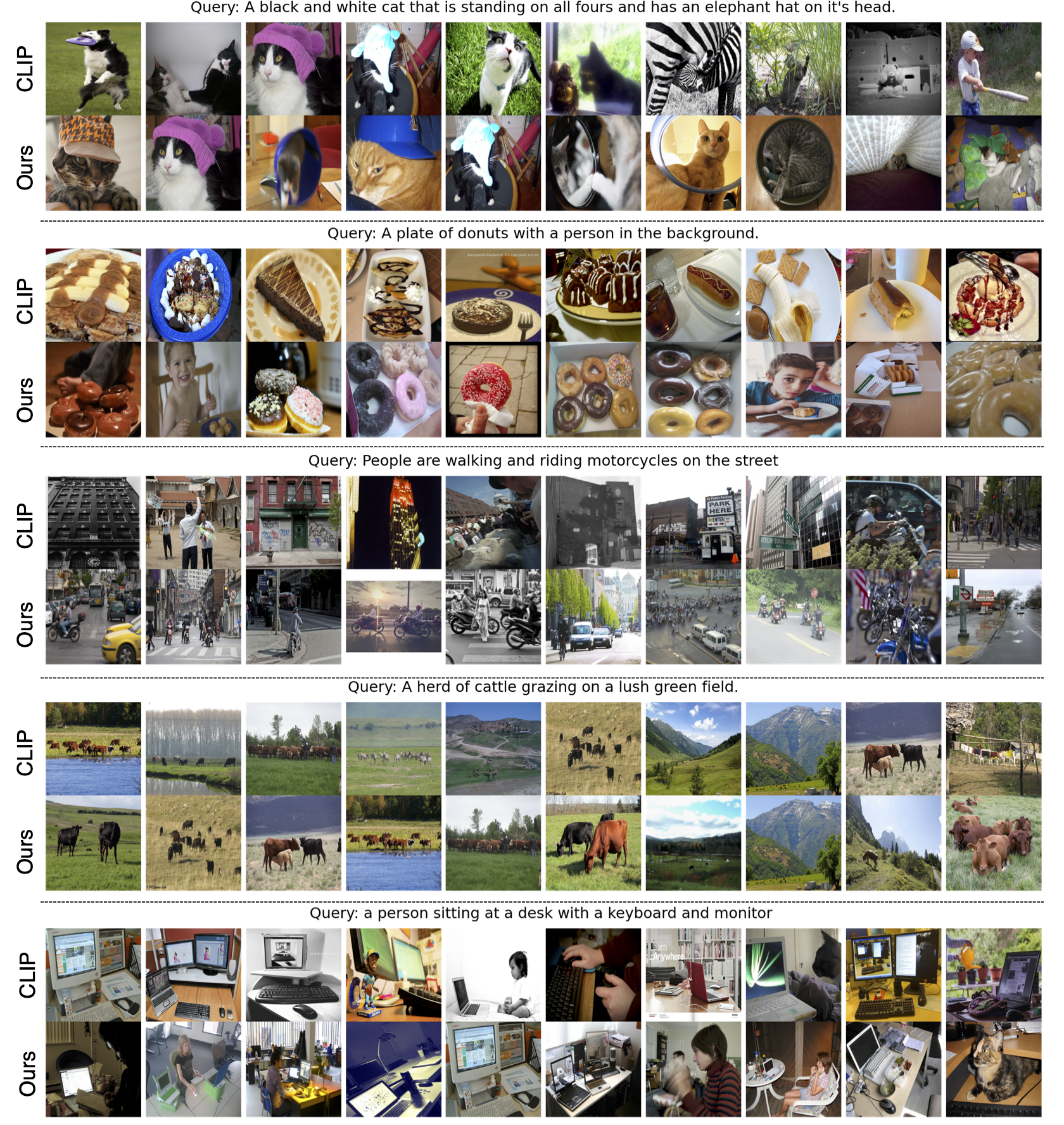}
    \caption{
    \small
    \textbf{Example Text-Image Retrievals --} Given a text query, we display the top ten most semantically related images (ranked left to right) retrieved by CLIP and our method. Compared to CLIP, our method continues to retrieve images that more holistically match the text description, even after the ground truth image has appeared in the ranking.
    }
    \label{fig:qual_large}
\end{figure*}

\begin{figure*}[ht!]
    \centering
    \includegraphics[width=0.99\linewidth]{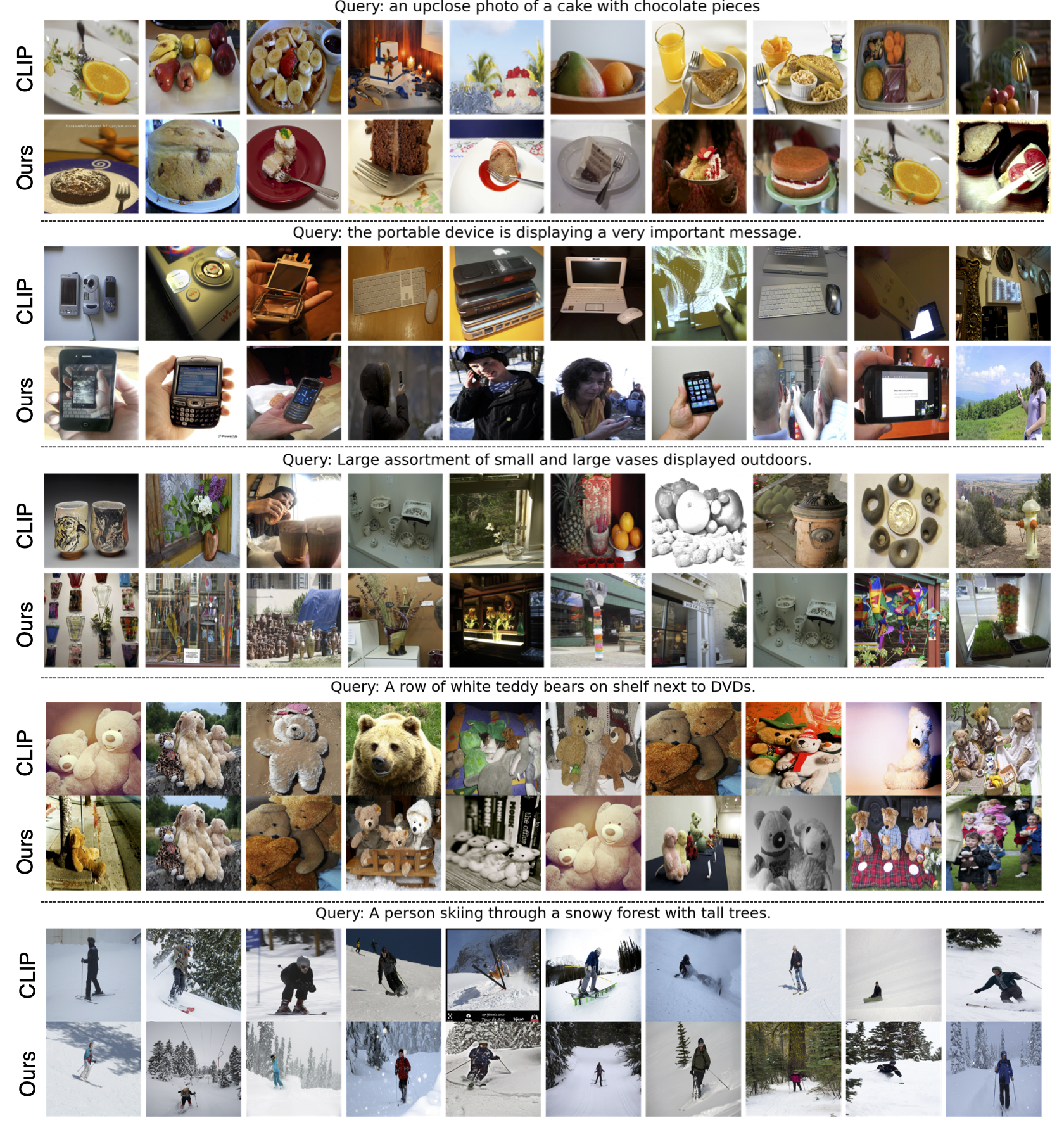}
    \caption{
    \small
    \textbf{Additional Text-Image Retrievals --} Given a text query, we display the top ten most semantically related images (ranked left to right) retrieved by CLIP and our method. Compared to CLIP, our method continues to retrieve images that more holistically match the text description, even after the ground truth image has appeared in the ranking. 
    }
    \label{fig:qual_supp}
\end{figure*}

\noindent
In Figures \ref{fig:qual_large} and \ref{fig:qual_supp}, we provide additional text-image retrieval results computed on the COCO test set. Consistent with the retrievals shown in Figure 6 from our main paper, our method more consistently captures the full semantic extent of the query caption, whereas the baseline CLIP model tends to narrowly focus on one particular aspect. For example, in the first row of Figure~\ref{fig:qual}, CLIP does retrieve a ``black and white cat,'' but also retrieves a black and white dog, a black and white zebra and a black and white photograph of boxes, whereas our method retrieves images containing a cat in 9 out of the 10 top retrievals.

\subsection{Additional Robust Classification Examples}
\begin{figure*}[ht!]
    \centering
    \includegraphics[width=0.75\linewidth]{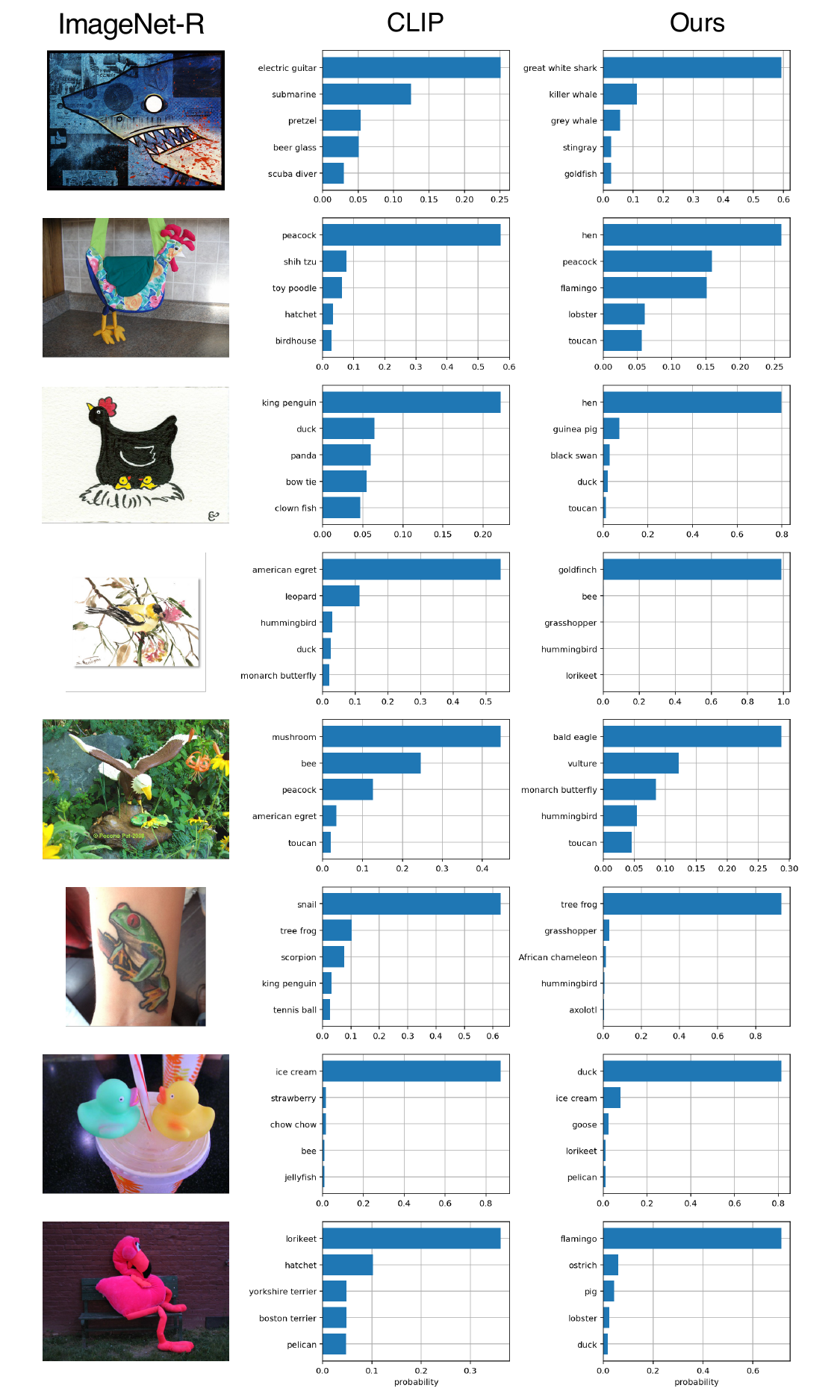}
    \caption{
    \small
    \textbf{Robustness Examples --} Given an image from the ImageNet-R dataset (left column), we compare the predictions of CLIP (middle column) to the predictions of our method (right column) by showing the probabilities assigned to the top 5 classes.
    }
    \label{fig:robustness_supp}
\end{figure*}

\noindent
In Figure~\ref{fig:robustness_supp}, we show qualitative examples of the instances where our method shows improved robustness over its CLIP counterpart for out-of-distribution images drawn from the ImageNet-R dataset. While CLIP struggles to handle certain artistic styles (angular shark depicted in row 1, goldfinch in row 4, tattooed tree frog in row 6), strong color patterns (black and white colors in row 3), or subjects in unusual contexts (rows 5,7,8), our method is able to more consistently provide a reasonable set of top predictions, which is consistent with the quantitative improvement of nearly 8.5\% on average when considering Conceptual Captions pretraining as reported in Table 1 of our main paper.

\subsection{Similarity Matrix Visualization}
\begin{figure*}[t!]
    \centering
    \includegraphics[width=1.0\linewidth]{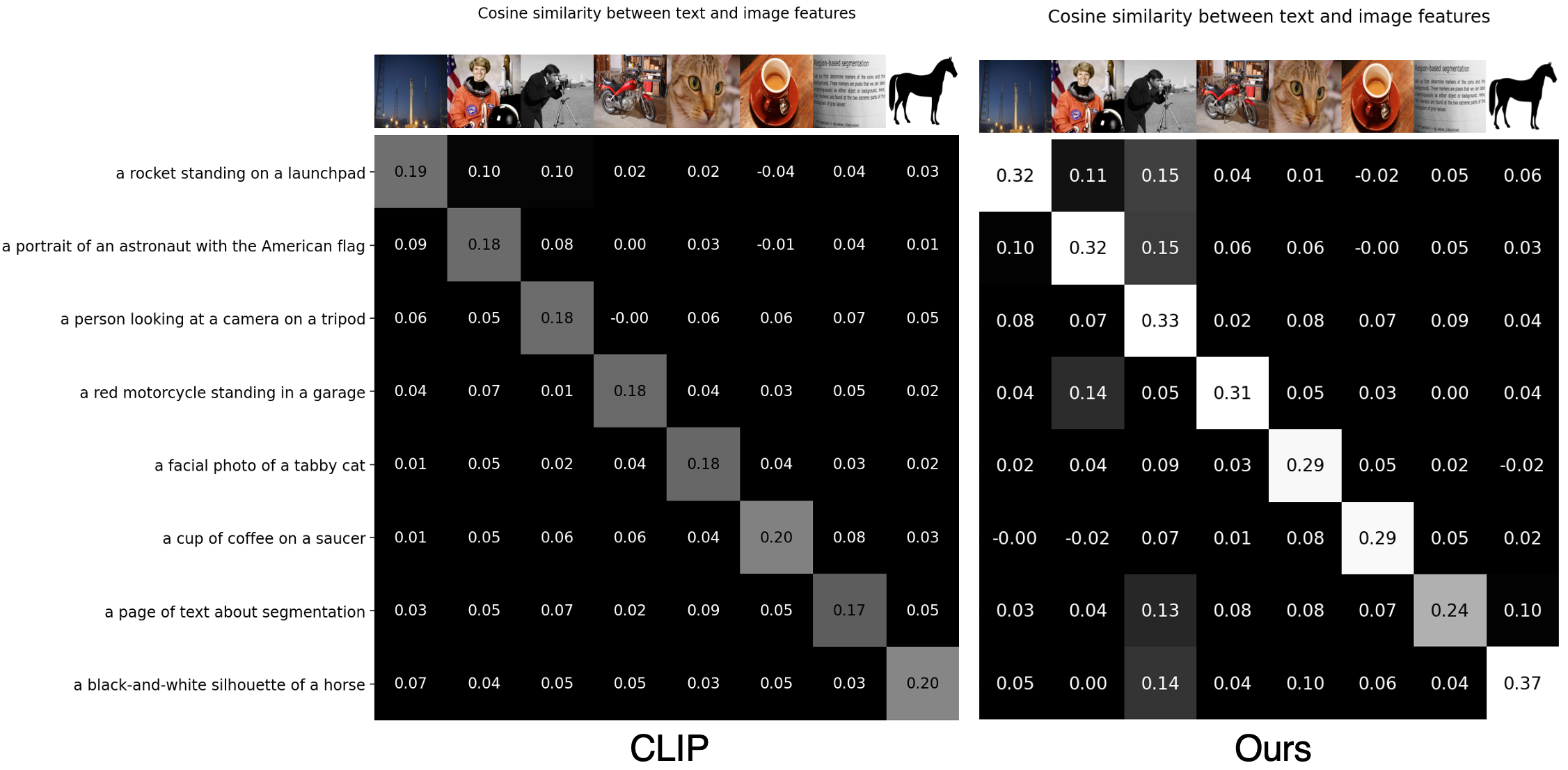}
    \caption{
    Visualization of our method's similarity scores between a batch of eight image-text pairs.
    The baseline CLIP is optimized to maximize the diagonal scores and minimize off-diagonal scores, even when there exists non-negligible semantic similarity between unpaired instances. In contrast, our method yields elevated similarity scores on off-diagonal elements when there is increased semantic similarity between unpaired instances (e.g., photographer-to-astronaut pairing and the black-and-white-to-photographer-to-page of text pairing).
    As a result, we empirically observe larger similarity scores for ground truth positive pairs with our method, which coincides with improved downstream zero-shot classification performance.
    }
    \label{fig:similarity_vis}
\end{figure*}

\noindent
In Figure~\ref{fig:similarity_vis}, we show a matrix of pairwise cosine similarity scores assigned to a batch of images and corresponding text snippets by our method compared to its CLIP counterpart.
These similarity matrices present concrete examples of the trends captured by the distributions shown in Figure 5 of the main paper.
Namely, it shows that CLIP has been optimized to assign high similarity scores along the diagonal (positive pairs) and low similarity to off diagonal elements (negatives), even when there is non-negligible semantic similarity between unpaired instances (\emph{e.g.,} the text ``a black-and-white silhouette...'' and the black-and-white image of a photographer dressed in black clothing) .
In contrast, our method yields elevated scores for negative pairings that show this amount of secondary similarity.
As a consequence, we empirically observe that our learned representations produce larger scores for ground truth positive pairs and lead to more robust zero-shot classification performance. 

\section{Ethical Considerations}

\subsection{Impact on ML and Related Scientific Fields}
\noindent
A primary motivation driving this work is to increase the robustness and efficiency of a vision-language pretraining (VLP) method that has recently given rise to a set of so-called \emph{foundation models}~\cite{bommasani2021opportunities}.
Due to the anticipated role that foundation models are to play in the immediate development of AI systems, contributions to advancing the core training method will have far-reaching impacts on the field and downstream application areas by definition.
Since improving the efficiency and lowering the computational/environmental cost associated with this VLP method is a primary objective of our work, we would like our work to assist in providing greater accessibility to the study, development and deployment of these VLP methods.

\subsection{Impact on Society}
\noindent
Robustness to challenging, novel, and even adversarial examples is rapidly becoming an extremely important part of modern computer vision systems, which are now starting to be deployed in sensitive contexts such as autonomous vehicles~\cite{ingle2016tesla} and medical applications~\cite{davenport2019potential} with life and death consequences.
Additionally, the increasing diversity of data sources, ranging from massive and cumbersome datasets to extremely limited and highly sensitive information, poses several practical and environmental challenges to consistently training robust and reliable machine learning systems. Our proposed framework aims to address these aspects jointly.

{\small
\bibliographystyle{ieee_fullname}
\bibliography{supp}
}